\documentclass[11pt, letterpaper, shortlabels]{berkeley}

\theoremstyle{plain}

\theoremstyle{definition}

\theoremstyle{remark}

\usepackage[numbers]{natbib}
\usepackage{enumitem}

\usepackage{wrapfig}
\usepackage{graphicx}
\usepackage{subcaption}
\usepackage{fontawesome5}
\usepackage{diagbox}

\usepackage{algorithm}
\usepackage{algpseudocode}
\usepackage{booktabs}
\usepackage{colortbl}

\usepackage{siunitx}
\usepackage{multirow}
\usepackage{microtype}
\usepackage{tabu}

\newcolumntype{M}[1]{>{\centering\arraybackslash}m{#1}}
\renewcommand{\arraystretch}{1.2}

\usepackage{pifont}
\usepackage{adjustbox}
\usepackage{xspace}

\usepackage{enumitem}

\usepackage{pifont}

\newcommand{\cmark}{\textcolor{ForestGreen}{\ding{51}}}
\newcommand{\xmark}{\textcolor{BrickRed}{\ding{55}}}

\definecolor{BrickRed}{rgb}{0.8, 0.25, 0.33}
\definecolor{ForestGreen}{rgb}{0.13, 0.55, 0.13}

\usepackage{tikz}
\newcommand*\circledblue[1]{\tikz[baseline=(char.base)]{
            \node[shape=circle,draw=NavyBlue!100,fill=NavyBlue!10,thick,inner sep=1pt] (char) {\scriptsize\textsf#1};}}

\newcommand*\circledred[1]{\tikz[baseline=(char.base)]{
            \node[shape=circle,draw=BrickRed!60,fill=BrickRed!10,thick,inner sep=1pt] (char) {\scriptsize\textsf#1};}}

\definecolor{NavyBlue}{RGB}{0,127,255} 
\definecolor{myyellow}{RGB}{255,217,102} 

\definecolor{HeaderMintPurple}{HTML}{EEEAF7}
\definecolor{RowMintPurple}{HTML}{F6F4FB}
\definecolor{TextDark}{HTML}{2B2B2B}

\definecolor{IconChat}{HTML}{10B981}   
\definecolor{IconMath}{HTML}{3B82F6}   
\definecolor{IconCode}{HTML}{F59E0B}   
\definecolor{IconChem}{HTML}{A855F7}   
\definecolor{IconQA}{HTML}{EF4444}     
\definecolor{IconEval}{HTML}{06B6D4}

\newcommand{\LATER}[1]{}

\usepackage{fontawesome5}

\newcommand{\Methodlong}{\emph{Personalized Prompt Optimization}\xspace}
\newcommand{\methodshort}{PPOpt\xspace}
\newcommand{\dataset}{\textsc{PersonaAtlas}\xspace}
\newcommand{\dataGeneration}{\textsc{PersonaGym}\xspace}
\usepackage[most]{tcolorbox}
\usepackage{xcolor}
\hypersetup{
  colorlinks=true,
  linkcolor=violet!70!black,
  citecolor=violet!70!black,
  urlcolor=violet!70!black
}
\usepackage{array}
\newcolumntype{M}{>{$}c<{$}}  
\newcolumntype{L}{>{$}l<{$}}  
\newcolumntype{R}{>{$}r<{$}}

\newtcolorbox{motivationbox}{
  enhanced,
  colback=blue!3,
  colframe=blue!35,
  boxrule=0.5pt,
  arc=2.2pt,
  left=8pt,right=8pt,top=9pt,bottom=7pt,
  before skip=6pt,
  after skip=8pt,
  drop shadow={black!20},
}

\title{Synthetic Interaction Data for Scalable Personalization in Large Language Models}
\author[1]{Yuchen Ma\textsuperscript{\textdagger}}
\author[2]{Yue Huang}
\author[3]{Wenjie Wang}
\author[2]{Xiaonan Luo}
\author[2]{Xiangliang Zhang}
\author[1]{Stefan Feuerriegel}
\correspondingauthor={Yuchen Ma (\href{mailto:yuchen.ma@lmu.de}{yuchen.ma@lmu.de})}

\affil[1]{Munich Center for Machine Learning \& LMU Munich}
\affil[2]{University of Notre Dame}
\affil[3]{Sichuan University}

\begin{abstract}
Personalized prompting offers large opportunities for deploying large language models (LLMs) to diverse users, yet existing prompt optimization methods primarily focus on task-level optimization, while largely overlooking user-specific preferences and latent constraints of individual users. This gap is primarily due to (i)~the absence of high-quality, privacy-sensitive data that capture personalized user–LLM interactions at scale, and (ii)~the lack of robust reward signals for individual preferences. To overcome existing data limitations, we introduce a high-fidelity synthetic data generation framework called \dataGeneration. Unlike prior work that treats personalization as static persona--preference pairs, \dataGeneration models a dynamic preference process via an agentic LLM system to simulate realistic preference behaviors and semantic-aware noise in order to generate personalized \emph{multi-turn} interaction trajectories. Using \dataGeneration, we release \dataset, a large-scale, high-quality, and diverse synthetic dataset of high-fidelity multi-turn personalized interaction trajectories that closely mirror real-world preference expression and noise patterns. We further propose \Methodlong (\methodshort), a scalable and model-agnostic framework that optimizes user prompts based on interaction histories without modifying the deployed LLM. \methodshort adopts a reason-then-optimize paradigm that infers an explicit user profile and conditions prompt rewriting on the user profile to avoid rewarding hacking. Our training procedure for \methodshort integrates a cold-start supervised prior with outcome-driven multi-objective reinforcement learning. We present extensive experiments to demonstrate consistent improvements over state-of-the-art baselines in terms of task performance, personalization quality, and robustness to noisy as well as to sparse preference signals.
\end{abstract}

\begin{document}
\maketitle

\begin{center}
\vspace{0.3em}
{\setlength{\fboxsep}{6pt}%
\fcolorbox{BerkeleyBlue}{blue!3}{%
  \parbox{0.86\linewidth}{%
    \centering\textbf{\large Accepted at KDD 2026 (Datasets \& Benchmarks Track)}\par
  }%
}}
\vspace{0.3em}
\end{center}

\noindent
\faLink\ \textbf{Dataset:} \url{https://huggingface.co/datasets/HowieHwong/PPOpt-data}\\
\faCube\ \textbf{Model:} \url{https://huggingface.co/HowieHwong/ppopt}\\
\faBook\ \textbf{Docs:} \url{https://personagym.readthedocs.io/}\\
\faGithub\ \textbf{Code:} \url{https://github.com/yccm/LLM-PPOpt}

\section{Introduction}

Large language models (LLMs) are increasingly deployed as general-purpose assistants for writing, coding, planning, and decision support \citep{bai2022constitutional, brown2020language, Bu2025PersonalizedLD, Chen2024PADPA, Garbacea2025HyPerAlignIP, lee2023rlaif, rafailov2023direct, Tan2025AligningLL, Zhang2025PersonajudgePA, Zhao2025NextQuillCP}. Despite impressive general capabilities, real-world usage is inherently driven by \emph{personalized} needs: users may prefer different levels of verbosity and structure, different reasoning styles,  and different degrees of proactivity, and many of these preferences come as latent constraints that are seldom stated explicitly (such as, e.g., privacy concerns,  formatting requirements, domain-specific conventions) \citep{zhou2023instruction}. As a result, personalized alignment is important for improving users' experiences during LLM interactions.

However, current efforts to personalize alignment are limited due to a severe \circledred{1}~\emph{data gap} for two practical reasons: First, high-fidelity data of personalized multi-turn user interactions with LLMs are absent, because collecting and releasing such data is challenging due to privacy and consent reasons \citep{tran2025understanding}. Second, real user feedback is inherently noisy and incomplete: users rarely provide explicit rewards, and their responses can be ambiguous or inconsistent across turns. 

Moreover, existing approaches for personalized alignment have also \circledred{2}~\emph{method limitations} in that these approach typically rely on \emph{static} preferences, \emph{limited} feedback, or \emph{LLM-side} adaptation \cite{Bu2025PersonalizedLD, Chen2024PADPA, Garbacea2025HyPerAlignIP,Kim2025CUPIDEP,Tan2025AligningLL, Zhang2025PersonajudgePA, Zhao2025NextQuillCP,Zhang2025PROPERAP}. However, these approaches are often computationally expensive or inapplicable to proprietary, closed-source LLMs. Deployed LLMs are often proprietary or frozen, limiting access to parameters and gradients. These challenges limit the scalability of personalized alignment through model-side training. A natural way to personalize \emph{without} modifying the underlying model is \emph{prompt adaption}. Black-box prompt optimization methods can steer a fixed LLM by rewriting user instructions, providing a model-agnostic and potentially transferable mechanism for behavior control \cite{cheng2024black, Cui2024SEESE,Guo2023EvoPromptCL,wang2025adareasoner,Zhuravlev2025AutomaticPO}, which provides a promising way to solve this problem.

To address the above bottlenecks (i.e., \circledred{1}~\emph{data gap} and \circledred{2}~\emph{method limitation}) behind personalized alignment, we first introduce a \circledblue{1}\,high-fidelity synthetic data generation framework called \dataGeneration, and we further also present a \circledblue{2}\,model-agnostic framework for \Methodlong (\methodshort).

\circledblue{1}\,\textbf{Data generation framework (\dataGeneration):} In practice, user preferences are often expressed implicitly through interaction history: users interact with the LLM over multiple turns which allow users to encode preferences by correcting the LLM, rephrasing requests, and accepting or rejecting suggestions in order to gradually steer the LLM toward a preferred format for the output. A key novelty in \dataGeneration is that we introduce a scalable synthetic data generation pipeline that models each user as a \emph{dynamic latent preference process} rather than a static persona. Our \dataGeneration follows an agentic framework involving three LLMs (User, Assistant, and Distractor) to simulate realistic preference behaviors and semantic-aware noise. Concretely, we generate diverse multi-turn interaction trajectories where (1)~user intent can be partially specified, (2)~user updates their behavior in response to the LLMs's outputs, and (3)~noise is explicitly injected to mimic real conversational patterns. This synthetic construction differs from \textit{static} persona-preference pairs by treating conversational dialogues as an interactive, \textit{dynamic} process. Based on this, we release \dataset, a large-scale, high-fidelity synthetic dataset of high-fidelity multi-turn personalized interaction trajectories, generated by \dataGeneration to closely mirror real-world preference expression and noise patterns.

\circledblue{2}\,\textbf{\Methodlong (\methodshort):} 
We propose \Methodlong{} (\methodshort), a model-agnostic framework for \emph{scalable personalized prompt optimization} that decouples personalization from LLM training. \methodshort learns a \emph{personalized prompt optimizer} that rewrites user prompts based on interaction history, without updating the parameters of the deployed LLM. This black-box formulation is compatible with proprietary models and easy to deploy. \methodshort follows a \emph{reason-then-optimize} paradigm: it first infers a latent user profile from historical interactions, then rewrites the current prompt conditioned on the inferred profile. The optimizer is trained with \emph{task-level outcome rewards} derived from downstream LLM performance, yielding an outcome-driven objective that mitigates shortcut learning and improves generalization. We further adopt a multi-objective RL formulation to balance \emph{personalization fidelity} and task performance. We evaluate \methodshort on both synthetic benchmarks and real-world user data. Across settings, it consistently improves task outcomes, personalization quality, outperforming state-of-the-art baselines.

Overall, our \textbf{main contributions} are the following:
\begin{enumerate}[leftmargin=15pt]
\item \textbf{High-fidelity synthetic data generation framework for personalized interaction}: We introduce \dataGeneration, a high-fidelity synthetic data generation framework. Based on this, we release \dataset, a large-scale, diverse synthetic dataset of high-fidelity multi-turn personalized interactions.
\item \textbf{Prompt optimization method for personalization}: By leveraging \dataset, we propose a scalable and model-agnostic framework \methodshort that optimizes user prompts from interaction histories without modifying the deployed LLM. 
\item \textbf{Empirical validation}: We conduct extensive evaluations using the synthetic data generated by \dataGeneration to demonstrate the effectiveness of \methodshort. In particular, our \methodshort shows improved personalization and generalization to unseen users.
\end{enumerate}

\definecolor{ppoptbg}{RGB}{255, 243, 204}

\section{Related Work}
\label{sec:related_work}

\subsection{Synthetic Data Generation with LLMs}
LLMs have demonstrated strong capability in generating high-quality synthetic data at scale~\citep{liu2024best}. Compared to earlier approaches based on conventional language models~\citep{schick2021generating}, modern LLMs are highly expressive and controllable. Prior work has shown the effectiveness of LLM-generated data across various settings, including multilingual question answering~\citep{riabi-etal-2021-synthetic}, conversational systems~\citep{zhao2023inthe}, instruction tuning~\citep{xu2024magpie, zhang2025oasis, zhong2024synthet2c}, factuality improvement~\citep{wei2023simple}, synthetic data for scientific domains~\citep{huang2025chemorch, 10.1145/3711896.3737432}, and dataset diversification~\citep{dai2025auggpt, chung2023increasing, riaz2025metasynth}. Recent frameworks such as DataGen~\citep{huang2025datagen} and Janus~\citep{lee2024aligning} further highlight the scalability and versatility of LLM-based approaches for synthetic data generation. Given the strong generative capacity of LLMs, applying synthetic data generation to personalized conversations and interaction-level settings is particularly promising \citep{Li2025From1U, Chan2024ScalingSD, Wu2024AligningLW}. 

In contrast to prior works that treat personalization as static \emph{persona-preference} pairs, we model personalization as an \emph{dynamic} conversational process: preferences emerge only after observing the LLM’s response, thus our modeling approach closely reflects real-world user interactions and feedback (we later calls this also ``outcome-driven''). We further incorporate realistic human preference behaviors such as ambiguous intents and cross-turn inconsistency into our data generation process, thereby enabling training under noisy conditions that mirror real conversations.

\begin{table*}[t]
\centering
\small

\setlength{\tabcolsep}{3pt}
\renewcommand{\arraystretch}{1.15}
\caption{
Comparison of personalization methods. Shown are the following key dimensions:
\emph{Intervention level} indicates the stage at which personalization is applied (e.g., prompt-, memory-, decoding-, or parameter-level).
\emph{Data scalability} reflects whether the personalization data can be scaled, for example via synthetic data generation with a mature and reusable workflow.
\emph{User noise modeling} measures whether the method explicitly accounts for realistic human noise, such as ambiguous intent, typos, or inconsistent behaviors.
\emph{Interaction dynamics} denotes whether personalization is modeled over multi-turn interactions rather than single-turn inputs.
\emph{Closed-source compatibility} indicates whether the method can be applied to closed-source or proprietary models without accessing model parameters.
}
\label{tab:personalization_comparison}
\scalebox{0.67}{
\begin{tabular}{l|>{\columncolor{ppoptbg}}c cccccccc}
\toprule
\textbf{Dimension} 
& \textbf{PPOpt (\emph{ours})} 
& \textbf{LaMP \citep{salemi-etal-2024-lamp}} 
& \textbf{PERSONA AI \citep{Wang2024AIPT}} 
& \textbf{CAPM \citep{Pitis2024ImprovingCP}} 
& \textbf{Summ+Ret \citep{Richardson2023IntegratingSA}} 
& \textbf{Matryo. \citep{Li2024MatryoshkaPL}} 
& \textbf{PAD \citep{Chen2024PADPA}} 
& \textbf{ICPL \citep{Yu2024ICPLFI}} 
& \textbf{NextQuill \citep{Zhao2025NextQuillCP}} \\
\midrule

\textbf{Intervention level} 
& \textbf{Prompt} 
& Memory 
& Memory
& Parameter
& Memory
& Prompt
& Decoding
& Prompt 
& Parameter\\

\textbf{Data scalability} 
& \textbf{High} 
& High 
& High 
& Medium 
& High 
& Medium 
& Medium 
& Low 
& Medium \\
\midrule
\textbf{User noise modeling} 
& \textbf{\cmark} 
& \xmark 
& \xmark 
& \xmark 
& \xmark 
& \xmark 
& \xmark 
& \xmark 
& \xmark \\
\textbf{Interaction dynamics} 
& \textbf{\cmark} 
& \xmark 
& \cmark 
& \xmark 
& \xmark 
& \cmark
& \xmark 
& \xmark 
& \xmark \\
\textbf{Closed-source Compa.} 
& \textbf{\cmark} 
& \cmark 
& \cmark 
& \xmark 
& \cmark 
& \cmark 
& \xmark 
& \xmark 
& \xmark \\
\bottomrule
\end{tabular}}
\vspace{-5pt}
\end{table*}

\subsection{Prompt Optimization of LLMs}
A growing line of work studies \emph{black-box} prompt optimization, which rewrites user instructions without access to model parameters or gradients and which is thus applicable to proprietary or frozen LLMs. Early studies focus on single-prompt rewriting: BPO \citep{cheng2024black} automatically refines human-written prompts to elicit better behaviors from a fixed LLM, while DistillPrompt \citep{Zhuravlev2025AutomaticPO} distills task-specific knowledge from data into compact, high-performing prompts. Other methods search over prompts more explicitly; e.g., EvoPrompt \citep{Guo2023EvoPromptCL} uses evolutionary algorithms to evolve prompts, and SEE \citep{Cui2024SEESE} treats the target LLM as a black-box evaluator and optimizes prompts using only task-level feedback. Complementary to these, AdaReasoner \citep{wang2025adareasoner} targets model-agnostic improvements in reasoning via prompt optimization, and
Promptolution \citep{Zehle2025promptolutionAU} provides a modular framework that standardizes and unifies diverse prompt-optimization pipelines. 

Beyond rewriting a single instruction, recent work proposed methods to search over higher-level prompting \emph{patterns} and multi-step prompt content for LLM agents. Examples are AutoPDL \citep{Spiess2025AutoPDLAP}, AFlow \citep{Zhang2024AFlowAA}, and DyFlow \citep{wang2025dyflow}. Finally, self-supervised prompt optimization reduces the reliance on labeled data by leveraging an LLM’s own outputs as supervision (e.g., by comparing candidate prompts through self-consistency or preference signals \citep{Xiang2025SelfSupervisedPO,Gao2025ThePA}).

Overall, prompt optimization offers a powerful route to steering LLM behavior without training or parameter access. Prompt optimization is inherently model-agnostic and often transferable across different LLMs. However, despite these properties, prompt optimization has been relatively under-explored in the context of \emph{personalization}, where model-agnostic adaptation and transferability are particularly beneficial.

\vspace{-4pt}
\subsection{Personalized Alignment of LLMs}

A large body of work aligns LLMs with general human preferences via reinforcement learning and preference optimization, including reinforcement learning from human feedback (RLHF) \cite{ouyang2022training}, reinforcement learning from AI feedback (RLAIF) \cite{bai2022constitutional, lee2023rlaif}, and direct preference optimization (DPO) \cite{rafailov2023direct}. While these approaches are effective at modeling \textit{population-level} preferences, they struggle to capture \textit{individual-level} user heterogeneity, thus motivating \emph{personalized alignment} \citep{Guan2025ASO, Zhang2024PersonalizationOL}. 

Existing methods for personalized alignment can be broadly grouped by \textit{where} adaptation occurs. One line of research targets \emph{inference-time} personalization without per-user fine-tuning. PAD \citep{Chen2024PADPA} injects user-specific rewards into token-level decoding, CoPe \citep{Bu2025PersonalizedLD} contrasts user-specific preferences against a base model, and Persona-judge \citep{Zhang2025PersonajudgePA} uses the model’s own preference judgments to filter candidate tokens. A second line improves \emph{preference modeling} and data efficiency. NextQuill \citep{Zhao2025NextQuillCP} adopts a causal framework to isolate the effects of user history, HyPerAlign \citep{Garbacea2025HyPerAlignIP} infers interpretable preference hypotheses from a small number of examples, and PUGC \citep{Tan2025AligningLL} leverages implicit preferences from unlabeled user-generated content.
A third line studies \emph{long-term} (system-level) personalization. AI PERSONA \citep{Wang2024AIPT} enables lifelong personalization through continual user profile updates, while PROPER \citep{Zhang2025PROPERAP} bridges group-level and individual-level preferences via data-efficient adaptation. In parallel, several other architectural, parameter-efficient approaches adapt models via lightweight modules, including Profile-to-PEFT \citep{Tan2025InstantPL}, HYDRA \citep{Zhuang2024HYDRAMF}, and modular PEFT composition \citep{Tan2024PersonalizedPE}. 
Finally, CUPID \citep{Kim2025CUPIDEP} introduces a multi-turn benchmark for evaluating context-dependent personalization in interactive settings. PERSONAMEM-V2 \citep{jiang2025personamemv2} targets personalization where preferences are mostly \emph{implicitly} revealed across long interaction histories. For this, PERSONAMEM-V2 introduces a dataset and studies how models can infer user personas from extended context; it further proposes an {agentic memory} framework that distills long histories into a compact, human-readable memory representation to improve personalization efficiency.

However, existing personalized alignment methods typically rely on \emph{static} preferences, \emph{limited} feedback, or \emph{model-/memory-side} adaptation (e.g., parameter updates, specialized memory modules, or very long-context inference). In contrast, our approach treats personalization as a scalable bridge between users and models: we translate \textit{latent} user intents into model-effective prompts using synthetic interaction data and outcome-driven reinforcement learning, thus achieving personalization without modifying the deployed model. We show the detailed comparison in Table~\ref{tab:personalization_comparison}.

\textbf{Research gap:} As discussed above, there is---to the best of our knowledge---\textbf{\underline{no}} approach for prompt \textit{personalization} that learns prompt rewrites from \emph{multi-turn} interactions trajectories while explicitly modeling \emph{latent} user preferences. To fill this gap, we propose a scalable, outcome-driven, and model-agnostic framework that jointly infers user preferences from interaction history and learns to generate personalized prompt rewrites optimized for downstream outcomes.

\vspace{-7pt}
\section{\dataGeneration: High-Fidelity Personalized Synthetic Interaction Data Generator}
\label{sec:synth-data}

\begin{figure*}
    \centering
    \includegraphics[width=\linewidth]{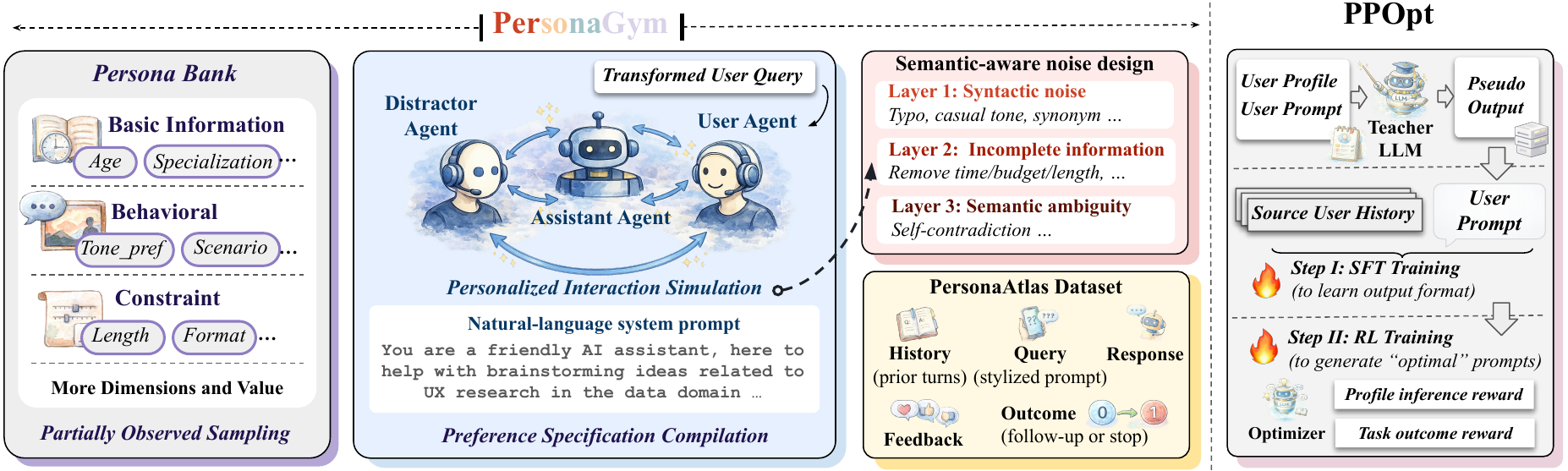}
    \vspace{5pt}
    \caption{Overview of our high-fidelity synthetic data generation framework \dataGeneration (left) and the unified prompt optimization framework \methodshort (right).
    }
    \label{fig:overview}
    \vspace{-8pt}
\end{figure*}

We propose \dataGeneration, a high-fidelity synthetic data generation pipeline that models users as dynamic preference processes (see Figure~\ref{fig:overview}, left). It generates diverse multi-turn interaction trajectories while capturing realistic user behaviors.

\vspace{-5pt}
\subsection{Overview of data generation}
\label{subsec:overview_data_gen}

Our synthetic data generator employs an agentic architecture, which involves three LLMs: (1)~a \textsc{User} model that generates user-like queries and feedback, (2)~an \textsc{Assistant} model that produces responses to mimic human preferences, and (3)~a \textsc{Distractor} model that injects realistic noise into both initial queries and follow-up feedback. The noise helps generate more human-like text (e.g., it includes spelling errors), which, in turn, improves robustness in downstream learning.
For this, our data generation pipeline operates in two stages:
(1) \textbf{persona-to-policy compilation}, where a persona is sampled from a curated \emph{Persona Bank}, optionally masked to model partial observability, and compiled into a system-level preference specification; and
(2) \textbf{trajectory synthesis}, where stylized initial queries are generated, multi-turn query–response–feedback interactions are simulated (using the \textsc{User} and \textsc{Response} agent), and trajectories are optionally perturbed (using the \textsc{Distractor} agent).
\begin{motivationbox}
\paragraph{Rationale for the agentic LLM approach in \dataGeneration:}
We employ different LLMs in a multi-agent setup---including the \textsc{User}, \textsc{Assistant}, and \textsc{Distractor} agent---for two reasons.
First, prior work~\cite{wu2025collabllm} has shown that instruction-tuned LLMs can serve as effective simulators of user behaviors.
Second, using separate LLMs disentangles the different sources of variation in interactive data: the \textsc{User} agent controls goal- and preference-conditioned behavior; the \textsc{Assistant} agent provides responses whose quality can be evaluated; and the \textsc{Distractor} agent injects controlled stochasticity that mimics real-world interaction noise.
\end{motivationbox}

\subsection{Persona Bank and Preference Specification Compilation}
\label{subsec:persona-bank}

\paragraph{Persona Bank.}
We construct a \emph{Persona Bank} $\mathcal{P}$ that captures heterogeneous user characteristics, preferences, and latent constraints. Each persona $p \in \mathcal{P}$ is represented as a set of feature--value pairs $p = \{(f_i, v_i)\}_{i=1}^{d}$, where $f_i$ denotes a feature dimension and $v_i$ its associated categorical value, as described in the following.

The feature dimensions span three schema-level categories in our persona specification: \textit{(i) basic user profile attributes}, \textit{(ii) interaction and behavioral traits}, and \textit{(iii) output/response constraints} (the details of these categories are shown in Appendix~\ref{app:details_data_generation}).
Moreover, we explicitly tag each dimension with an \texttt{is$\_$constraint} indicator to distinguish hard constraints, which must be strictly satisfied during generation (e.g., core profile attributes and response-level requirements), from soft constraints, which are treated as optional or partially observed signals (e.g., preferences for tone). This design enables controllable persona instantiation under partial observability to mimic realistic diversity in user intent and interaction dynamics while maintaining a compact categorical representation suitable for large-scale synthetic trajectory generation.

\paragraph{Feature sampling (under partial observability).}
In practice, user features are rarely fully observable. To mirror that such user features are only partially observable from the past history in real-world settings, we make information about the user ``sparse'' by employing stochastic sampling. Specifically, we model this by sampling an observed \emph{subset} of features
\vspace{-4pt}
\begin{equation}
\label{eq:obs-features}
o \sim \mathrm{Sample}(p; \pi_{\mathrm{mask}}), \quad o \subseteq p,
\end{equation}

where $\pi_{\mathrm{mask}}$ is a masking policy (e.g., random masking per feature). This yields a partially observed persona $o$.

Given the observed persona $o$, we compile a system-level preference specification $s$ that can condition the user simulation:
\begin{equation}
\label{eq:compile}
s = \mathrm{Compile}(o; \Psi),
\end{equation}

where $\Psi$ is an LLM-based \emph{Preference Spec Compiler} that translates structured attributes into a natural-language system prompt (details are in Appendix~\ref{app:implementation_details} and in our code).

\subsection{Personalized Interaction Simulation}
\label{subsec:trajectory}

We synthesize multi-turn interaction trajectories via an interaction generation pipeline. It employs three specialized stages to simulate realistic user-assistant dialogues:

\paragraph{(i) User query transformation}

We initialize the process using a seed query $q$ sampled from a mixture of public task datasets $\mathcal{D}$ (e.g., QA, rewriting, coding, planning; see Table~\ref{tab:dataset_query_domain}), denoted as $q \sim \mathcal{D}$. To induce user-specific phrasing, we optionally transform $q$ into a persona-stylized query $\tilde{q}$:

\begin{equation}
\label{eq:stylize}
\tilde{q} =
\begin{cases}
\mathrm{Stylize}(q, s; M_{\mathrm{user}}), & \text{with probability } \rho,\\
q, & \text{with probability } 1-\rho,
\end{cases}
\end{equation}

where $M_{\mathrm{user}}$ refers to the \textsc{User} agent l and $\rho$ controls how often queries are explicitly persona-stylized (here we set $\rho = 0.5$). We show an illustrative example of this process in Appendix~\ref{app:details_data_generation}.

\begin{motivationbox}
\paragraph{Why is the stylization stochastic?}
We deliberately do \emph{not} stylize every seed query for two reasons.
\textbf{(i)}~\emph{Intermittent signals.}
In practice, users alternate between plain task-only requests and preference-revealing formulations, so stylizing all queries would unrealistically densify personalization cues.
\textbf{(ii)}~\emph{Avoiding biased optimization.}
Over-stylized histories can bias the optimizer toward the initial query and reduce reliance on preference signals expressed in later feedback. As a result, stochastic stylization mitigates this by mixing clean and persona-tinted queries, encouraging preference recovery under partial observability.
\end{motivationbox}

\paragraph{(ii) Multi-turn interaction generation.}
We then simulate a conversation of up to $T$ rounds. At each round $t$:
the \textsc{Assistant} agent $M_{\mathrm{asst}}$ produces an answer $a_t$ given the current user message and history $h_t$;
the \textsc{User} agent emits feedback $f_t$ conditioned on a persona specification $s$, the response of the \textsc{Assistant} agent, and the query:
\vspace{-5pt}
\begin{align}
a_t &= M_{\mathrm{asst}}(\tilde{q}_t, h_{t-1}),\\
f_t &= M_{\mathrm{user}}(s, \tilde{q}_t, a_t, h_{t-1}).
\end{align}
In our simulation, if the \textsc{User} agent is not satisfied with the response, the agent produces a follow-up request $\tilde{q}_{t+1}$; otherwise, the trajectory terminates. This process yields a trajectory $\mathcal{T}_u=\left\{\left(q_1, a_1, f_1\right), \ldots,\left(q_T, a_T\right)\right\}$ for each user $u$.

\paragraph{(iii) Outcome-derived labels (positive/negative).}
We derive supervision signals for reinforcement learning (RL) from \emph{behavioral outcomes} (i.e., from the interaction trajectory) rather than from the prompt itself. To this end, if a user issues a follow-up request (or corrective feedback) after receiving $a_t$, we treat the preceding prompt instance as a \emph{negative} outcome; if the interaction terminates without follow-up, we treat it as \emph{positive}.
Let $\mathbb{I}[\cdot]$ denote an indicator, we define a binary label for turn $t$:
\begin{equation}
\label{eq:label}
y_t = 1 - \mathbb{I}[\text{follow-up at } t],
\end{equation}
where $y_t=1$ indicates satisfaction. We store both (i) the \emph{historical prompt trajectory} and (ii) turn-level positive/negative instances for downstream RL training.

\subsection{Data Augmentation via Distractor}
\label{subsec:distractor}

Real-world user prompts and feedback are often noisy, underspecified, or internally inconsistent (e.g., typos, missing constraints, vague references, or preference changes).
Relying solely on clean, fully specified synthetic interactions can therefore lead to unrealistic assumptions about user behavior.
To better capture the long tail of realistic inputs observed in practice, we simulate such variability explicitly.
We thus introduce a semantic-aware \textsc{Distractor} agent $M_{\mathrm{dist}}$ that injects controlled noise into both the initial user query and the follow-up feedback.

The \textsc{Distractor} agent acts as a perturbation engine that injects (a)~\emph{initial user query perturbation} and (b)~\emph{follow-up feedback perturbation}. Formally, given a clean persona-stylized query $\tilde{q}$ and feedback $f_t$, we generate corrupted versions via:
\begin{align}
\tilde{q}' &= \mathrm{Noisify}(\tilde{q}; M_{\mathrm{dist}}, \alpha_q),\\
f_t' &= \mathrm{Noisify}(f_t; M_{\mathrm{dist}}, \alpha_f),
\end{align}
where $\alpha_q,\alpha_f$ control overall noise strength. We then replace $\tilde{q}$ and/or $f_t$ with $\tilde{q}'$ and/or $f_t'$ during the interaction simulation process, yielding a corrupted trajectory $\tilde{\mathcal{T}}_u$ that the optimizer must learn to interpret.

\paragraph{Three-layer semantic-aware noise design.}
The \textsc{Distractor} is configured as a \emph{three-layer} corruption process (see the example in Figure~\ref{fig:distractor}). Intuitively, the layers represent increasing semantic disruption:

\begin{itemize}[leftmargin=*]
    \item \textbf{Layer 1: Syntactic noise} (intent intact). Preserve semantic content but perturb surface form (casual tone, punctuation changes, mild typos, synonym rewrite).
    \item \textbf{Layer 2: Incomplete information} (intent clear, information missing/vague). Preserve the core intent but drop or blur key cues/constraints (remove time/budget/length, replace with vague values, add context-dependence such as ``like last time'').
    \item \textbf{Layer 3: Semantic ambiguity} (intent uncertain or conflicting). Introduce ambiguity or contradictions (multi-intent concatenation, underspecified improvement requests, self-contradiction, mid-request mind changing).
\end{itemize}
The above layered design allows us to control how much the observed text deviates from the latent intent, matching realistic failure modes seen in deployed systems.

\subsection{Synthetic Data and \dataset\ Dataset}
\label{subsec:dataset}

We serialize each synthesized interaction into turn-level training instances following the trajectory construction in \S\ref{subsec:trajectory}--\S\ref{subsec:distractor}. For each persona $p\in\mathcal{P}$, we record partially observed features $o$ and compiled preference specifications $s$, unroll a (possibly corrupted) interaction trajectory, and extract one training instance per turn. A turn-level instance consists of the observed persona features, preference specification, interaction history, current (possibly distracted) user input and feedback, the assistant response, and a binary outcome label indicating whether the interaction terminates or continues.

By running \dataGeneration{}, we collect approximately 2{,}000 personas and over 10{,}000 conversations, which form the \dataset{} dataset. Example conversations are in the Appendix (see Figure~\ref{fig:PersonaAtlas_1}--Figure~\ref{fig:PersonaAtlas_4}.
). The dataset is publicly released on HuggingFace\footnote{\url{https://huggingface.co/datasets/HowieHwong/PPOpt-data}}.

\begin{table}[htbp]
\centering
\small
\setlength{\tabcolsep}{2pt}
\renewcommand{\arraystretch}{1}
\sisetup{
  group-separator = {,},
  group-minimum-digits = 4
}
\caption{Datasets used for query collection, with domains and query counts.}
\vspace{-3pt}
\label{tab:dataset_query_domain}

\resizebox{0.65\linewidth}{!}{%
\begin{tabular}{l l S[table-format=4.0]}
\toprule
\rowcolor{HeaderMintPurple}
\textcolor{TextDark}{\textbf{Dataset}} &
\textcolor{TextDark}{\textbf{Domain}} &
\textcolor{TextDark}{\textbf{\#Queries}} \\
\midrule

\rowcolor{RowMintPurple}
\texttt{ultrachat\_200k} \citep{ding2023enhancing} &
\textcolor{IconChat}{\faComments}\; Multi-turn chat / instruction &
3250 \\

\texttt{OpenR1-Math-220k} \citep{lozhkov2025openr1math220k} &
\textcolor{IconMath}{\faCalculator}\; Math reasoning &
2638 \\

\rowcolor{RowMintPurple}
\texttt{CodeFeedback} \citep{zheng2024opencodeinterpreter} &
\textcolor{IconCode}{\faCode}\; Code instruction \& feedback &
2028 \\

\texttt{Mol-Instructions} \citep{fang2023mol} &
\textcolor{IconChem}{\faFlask}\; Chemistry / molecules &
1766 \\

\rowcolor{RowMintPurple}
\texttt{MMLU} \citep{hendryckstest2021} &
\textcolor{IconQA}{\faGraduationCap}\; General knowledge (MCQ) &
1638 \\

\texttt{ai2\_arc} \citep{allenai:arc} &
\textcolor{IconQA}{\faLightbulb}\; Science QA (MCQ) &
1300 \\

\rowcolor{RowMintPurple}
\texttt{Alpaca-cleaned} \citep{alpaca} &
\textcolor{IconChat}{\faList}\; Instruction following &
844 \\

\texttt{TruthfulQA} \citep{lin2022truthfulqa} &
\textcolor{IconQA}{\faBalanceScale}\; Truthfulness / factual QA &
759 \\

\rowcolor{RowMintPurple}
\texttt{databricks-dolly-15k} \citep{DatabricksBlog2023DollyV2} &
\textcolor{IconChat}{\faRobot}\; Instruction / assistant data &
684 \\

\texttt{MBPP} \citep{austin2021program} &
\textcolor{IconCode}{\faTerminal}\; Programming problems &
619 \\

\rowcolor{RowMintPurple}
\texttt{MT-Bench} \citep{zheng2023judging} &
\textcolor{IconEval}{\faChartBar}\; Evaluation prompts &
80 \\
\bottomrule
\end{tabular}
} 
\vspace{-8pt}
\end{table}

\section{\Methodlong~(\methodshort)}
\label{sec:method}

Based on the large-scale synthetic dataset \dataset powered by \dataGeneration, we present \Methodlong (\methodshort), a unified framework that learns a \emph{personalized prompt optimizer} from user-specific interaction history. Given a user $u$ characterized by a synthetic persona $p_u$, historical trajectories $\mathcal{H}_u$, and a new session's initial query $\tilde{q}^{\mathrm{init}}_u$, \methodshort outputs an optimized prompt $\hat{q}^{\mathrm{init}}_u$ so that a \emph{fixed} base LLM produces responses that better satisfy the user’s personalization preferences.

\paragraph{Problem setup.} For each user $u$, we assume access to multiple historical sessions (each potentially multi-turn):
\begin{equation}
\mathcal{H}_u=\Big\{\mathcal{T}_u^{(j)}\Big\}_{j=1}^{J_u},
\qquad
\mathcal{T}_u^{(j)}=\Big\{(\tilde{q}^{(j)}_{t},a^{(j)}_{t},f^{(j)}_{t})\Big\}_{t=1}^{T_j},
\label{eq:ppopt_history}
\end{equation}
where $\tilde{q}^{(j)}_t$ denotes the user message, $a^{(j)}_t$ the assistant response from the
base assistant model $M_{\mathrm{asst}}$, and $f^{(j)}_t$ the user feedback.

In addition, we observe the \textbf{initial query} $\tilde{q}^{\mathrm{init}}_u$ of a new session.
The prompt optimizer receives the \emph{same input format during training and evaluation}, i.e., the initial query $\tilde{q}^{\mathrm{init}}_u$ with the historical trajectories $\mathcal{H}_u$:
\begin{equation}
s_u \;=\;\big(\tilde{q}^{\mathrm{init}}_u,\;\mathcal{H}_u\big).
\label{eq:ppopt_state}
\end{equation}

\subsection{Reason-Then-Optimize Paradigm}
\label{subsec:rto}

Directly mapping an interaction history to a rewritten prompt often encourages shortcut learning. To avoid this, we therefore design a \textbf{reason-then-optimize} paradigm: the optimizer first infers a natural-language user profile description,and then generates an improved prompt conditioned on it.

Formally, the optimizer is a conditional policy $\pi_\theta$ producing:
\begin{equation}
(\hat{z}_u,\;\hat{q}^{\mathrm{init}}_u)\sim\pi_\theta(\cdot\mid s_u),
\label{eq:ppopt_action}
\end{equation}
where $\hat{z}_u$ includes inferred user preferences/constraints/style based on $\mathcal{H}_u$.
We implement this with an autoregressive factorization:
\begin{equation}
\pi_\theta(\hat{z}_u,\hat{q}^{\mathrm{init}}_u \mid s_u)
=
\pi_\theta(\hat{z}_u\mid s_u)\cdot
\pi_\theta(\hat{q}^{\mathrm{init}}_u\mid s_u,\hat{z}_u),
\label{eq:ppopt_factor}
\end{equation}
realized by a single sequence model with explicit delimiters:
\[
\texttt{<REASONING>}~\hat{z}_u~\texttt{</REASONING>}~
\texttt{<PROMPT>}~\hat{q}^{\mathrm{init}}_u~\texttt{</PROMPT>}.
\]

\subsection{Cold-Start Supervised Fine-Tuning (SFT)}
\label{subsec:sft}

We use supervised fine-tuning (SFT) \emph{only} as a cold-start to (i) enforce the \textbf{reasoning-then-optimization output format} and (ii) teach the optimizer to \emph{condition} on $(\tilde{q}^{\mathrm{init}}_u,\mathcal{H}_u)$ to produce a plausible profile description followed by a rewritten prompt. Importantly, we do \emph{not} assume access to ground-truth ``optimal'' prompts; correctness and optimality are instead learned in the subsequent outcome-driven RL stage.

\paragraph{Pseudo targets via strong teacher synthesis.}
Let $\mathcal{D}_{\mathrm{train}}$ denote the synthetic dataset. Each training sample contains a state $s_u=(\tilde{q}^{\mathrm{init}}_u,\mathcal{H}_u)$ and the synthetic persona $p_u$ (or its observed subset), where $p_u$ is available \emph{only} during training. We use a powerful teacher model, i.e., a LLM (we use \texttt{GPT-5.2} for the implementation), to synthesize \emph{format-consistent} targets:
\begin{equation}
z^\star_u \sim \mathcal{T}_{\mathrm{prof}}(\cdot \mid p_u,\mathcal{H}_u),\qquad
q^\star_u \sim \mathcal{T}_{\mathrm{prompt}}(\cdot \mid \tilde{q}^{\mathrm{init}}_u,\mathcal{H}_u,z^\star_u),
\label{eq:pseudo_targets}
\end{equation}
where $\mathcal{T}_{\mathrm{prof}}$ produces a natural-language profile summary and $\mathcal{T}_{\mathrm{prompt}}$ produces a rewritten initial prompt. These pseudo labels
serve as \emph{scaffolding} to bootstrap generation structure; they are not treated as ground-truth optimal solutions. We show the synthesized case data in Appendix~\ref{app:case_study}. To reduce the noise within pseudo targets, we introduce a filtering mechanism, which is detailed in Appendix~\ref{app:implementation_details}.

\subsection{Outcome-Driven Multi-Objective RL}
\label{subsec:ppopt_rl}

After SFT, we optimize $\pi_\theta$ with RL using \textbf{downstream outcomes}. Given $(\hat{z}_u,\hat{q}^{\mathrm{init}}_u)$, we obtain the response from the \textsc{Assistant} agent under the improved prompt:
\begin{equation}
\hat{a}^{\mathrm{init}}_u = M_{\mathrm{asst}}(\hat{q}^{\mathrm{init}}_u;\mathcal{H}_u),
\label{eq:ppopt_hat_answer}
\end{equation}
and a baseline response under the original prompt:
\begin{equation}
a^{\mathrm{base}}_u = M_{\mathrm{asst}}(\tilde{q}^{\mathrm{init}}_u;\mathcal{H}_u).
\label{eq:ppopt_base_answer}
\end{equation}
(For a single-turn evaluation, $M_{\mathrm{asst}}(\cdot;\mathcal{H}_u)$ indicates that the historical context may be provided as additional conditioning text, but the optimized prompt is \emph{only} the new session’s first query.)

\paragraph{Multi-objective reward.}
We combine two rewards: (1) profile inference quality and (2) task outcome improvement:
\begin{equation}
R_u(\theta)
=
\lambda_{\mathrm{prof}}\,r_{\mathrm{prof}}(u)
+
\lambda_{\mathrm{task}}\,r_{\mathrm{task}}(u),
\qquad
\lambda_{\mathrm{prof}},\lambda_{\mathrm{task}}\ge 0.
\label{eq:ppopt_reward_sum}
\end{equation}

\paragraph{(1) Profile inference reward.}
In synthetic training, the ground-truth persona $p_u$ is available.
We score the inferred profile $\hat{z}_u$ with an LLM-as-judge $\mathcal{J}_{\mathrm{prof}}$:
\begin{equation}
r_{\mathrm{prof}}(u)
=
\mathrm{Norm}\Big(\mathcal{J}_{\mathrm{prof}}(\hat{z}_u,\;p_u)\Big)\in[0,1],
\label{eq:ppopt_r_prof}
\end{equation}
where $\mathcal{J}_{\mathrm{prof}}(\cdot)$ outputs a scalar rating (e.g., 1--10) for how well $\hat{z}_u$
matches the persona (including style and latent constraints), and $\mathrm{Norm}(\cdot)$ linearly rescales to $[0,1]$.

\paragraph{(2) Task outcome reward (pairwise, preference-aware).}
We evaluate whether the improved prompt yields a better assistant response than the original prompt.
We use an LLM-as-judge $\mathcal{J}_{\mathrm{task}}$ that is provided with:
the original query $\tilde{q}^{\mathrm{init}}_u$, the ground-truth persona $p_u$ (for preference grounding), and the two candidate responses $(a^{\mathrm{base}}_u,\hat{a}^{\mathrm{init}}_u)$.
The judge outputs a binary value reflecting whether the response using the improved prompt gives higher task correctness/clarity \textbf{and} preference satisfaction
\begin{equation}
r_{\mathrm{task}}(u)
=
\mathrm{Norm}\Big(
\mathcal{J}_{\mathrm{task}}(\tilde{q}^{\mathrm{init}}_u, p_u, a^{\mathrm{base}}_u, \hat{a}^{\mathrm{init}}_u)
\Big)\in \{0,1\}.
\label{eq:ppopt_r_task}
\end{equation}


As a result, this yields the trained \textit{optimizer} for personalizing LLMs. The optimizer $\pi_\theta$ is itself implemented as a language model, whose parameters are learned via SFT and outcome-driven RL Given a user’s interaction history and an initial query, this optimizer generates a rewritten prompt. When training the optimizer, we instantiate it using different base LLMs.

\section{Implementation details}
Due to the space limits, the implementation details are in Appendix~\ref{app:implementation_details}. We also release the code and toolkit docs for reference\footnote{Toolkit docs at \url{https://personagym.readthedocs.io/} and code is available at \url{https://github.com/yccm/LLM-PPOpt}}.

\section{Analysis of \dataset Dataset}

\paragraph{Token analysis.} 

Table~\ref{tab:token_breakdown} reports the average tokens per conversation sample used during the construction of \dataset. The results show that generating one conversation sample requires 17{,}662.8 output tokens in total, corresponding to \$0.2473/token for GPT-5.2 and \$0.1413/token for GPT-4.1.\footnote{Price estimates are based on information from OpenAI on February 1, 2026.} The interaction generation accounts for the vast majority of tokens (13{,}880; 78.6\%). These suggest that \dataGeneration is scalable for constructing large synthetic interaction datasets, as most cost is spent on producing diverse multi-turn interactions rather than repeatedly re-deriving persona specifications. This efficiency aligns with our goal of scalable personalization data generation.

\begin{table}[t]
\centering
\small
\setlength{\tabcolsep}{4pt}
\renewcommand{\arraystretch}{1}

\sisetup{
  group-separator = {,},
  group-minimum-digits = 4
}

\caption{Token consumption breakdown by different parts in \dataGeneration.}
\label{tab:token_breakdown}
\vspace{-3pt}

\resizebox{0.8\linewidth}{!}{%
\begin{tabular}{l S[table-format=5.1] S[table-format=3.1] S[table-format=1.4] S[table-format=1.4]}
\toprule
\rowcolor{HeaderMintPurple}
\textcolor{TextDark}{\textbf{Module}} &
\textcolor{TextDark}{\textbf{Tokens}} &
\textcolor{TextDark}{\textbf{Perc. (\%)}} &
\textcolor{TextDark}{\textbf{\$/token (GPT-5.2)}} &
\textcolor{TextDark}{\textbf{\$/token (GPT-4.1)}} \\
\midrule
\rowcolor{RowMintPurple}
Interaction generation & \bfseries 13880.0 & \bfseries 78.6 & \bfseries 0.1943 & \bfseries 0.1110 \\
Query generation       & 1824.6            & 10.3            & 0.0255           & 0.0146 \\
\rowcolor{RowMintPurple}
Distractor             & 1020.0            & 5.8             & 0.0143           & 0.0082 \\
Persona formulation    & 938.2             & 5.3             & 0.0131           & 0.0075 \\
\midrule
\rowcolor{RowMintPurple}
\textbf{Total}         & \bfseries 17662.8 & \bfseries 100.0 & \bfseries 0.2473 & \bfseries 0.1413 \\
\bottomrule
\end{tabular}
}
\end{table}

\begin{figure}[htbp]
    \vspace{-5pt}
    \centering
    \includegraphics[width=0.7\linewidth]{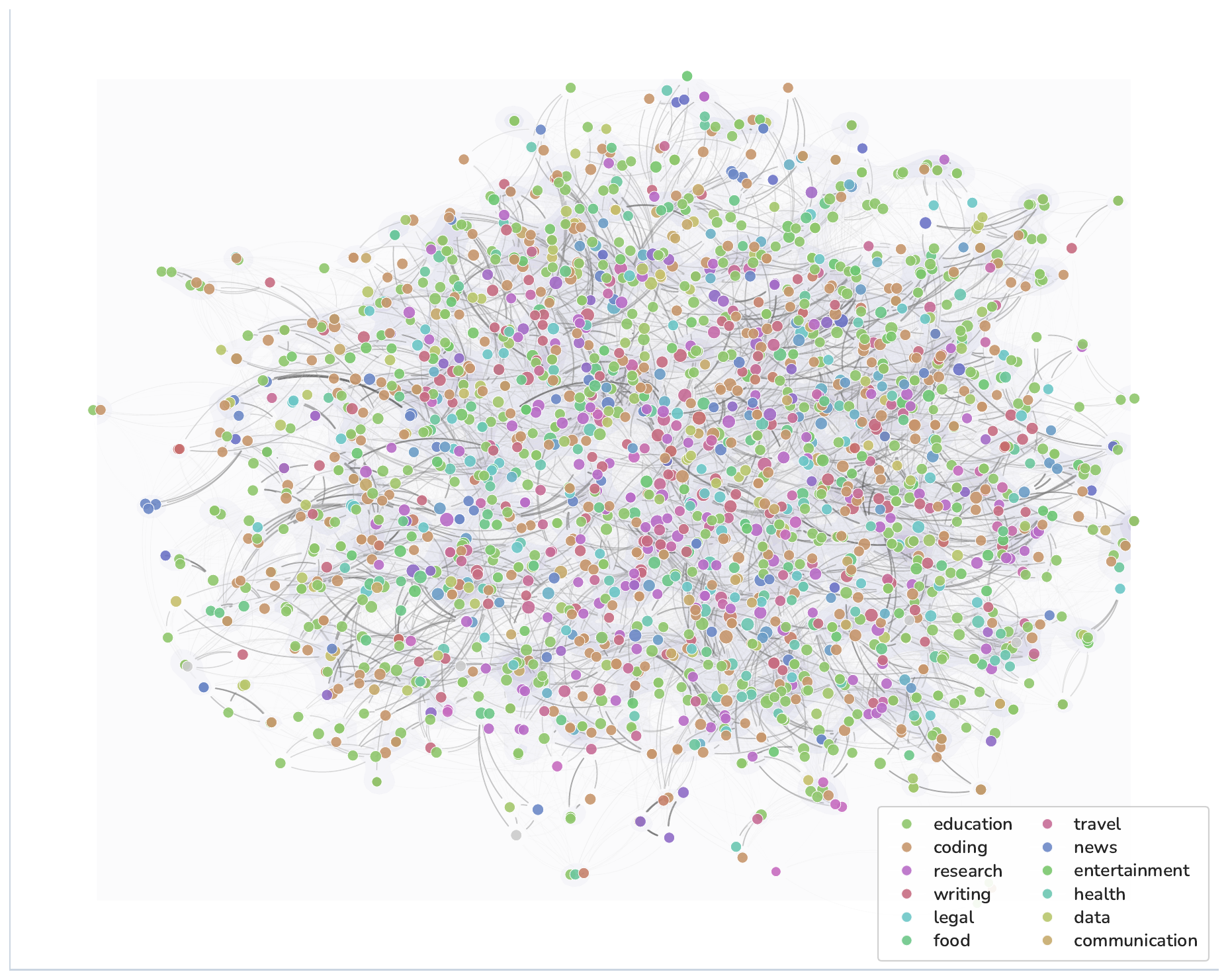}
    \caption{Conversation embedding of \dataset by different domains.}
    \label{fig:embedding}
\end{figure}

\paragraph{Embeddings.}

Figure~\ref{fig:embedding} visualizes conversation-level embeddings colored by domain. The data points are broadly interleaved rather than forming distinct clusters, thus indicating substantial semantic overlap across domains. This suggests that domain identity is not the dominant factor in the representation structure, which is intended by design.

\paragraph{Diversity.} Table~\ref{tab:diversity} compares data sample diversity across \textsc{UltraChat}~\cite{ding2023enhancing}, \textsc{ShareGPT}~\cite{sharegpt} (shareGPT52K), and ours. Self-BLEU measures intra-dataset similarity by computing BLEU of each sample against other samples; lower values indicate higher diversity. INGF reports the proportion of infrequent $n$-grams; higher values suggest more variable wording. The type--token ratio [TTR] (i.e., unique tokens divided by total tokens), captures lexical richness; higher is better. 

Our dataset achieves the lowest Self-BLEU, indicating substantially reduced intra-set repetition and higher sample-level variability. Our dataset also has the highest INGF and TTR, suggesting more frequent use of rarer lexical patterns and greater lexical richness overall. These results indicate that our data generation procedure produces a broader set of realistic realizations with less template-like overlap, which is desirable for training and evaluating models in realistic, human-like conversational phrasing.

\begin{table}[t]
\centering
\small
\setlength{\tabcolsep}{4pt}
\renewcommand{\arraystretch}{1}

\sisetup{
  group-separator = {,},
  group-minimum-digits = 4
}

\caption{Diversity comparison across datasets. Reported in Self-BLEU ($\downarrow$), INGF ($\uparrow$) and TTR ($\uparrow$).}
\label{tab:diversity}
\vspace{-4pt}

\begin{tabular}{l S[table-format=1.3] S[table-format=1.3] S[table-format=1.3]}
\toprule
\rowcolor{HeaderMintPurple}
\textcolor{TextDark}{\textbf{Dataset}} &
\textcolor{TextDark}{\textbf{Self-BLEU ($\downarrow$)}} &
\textcolor{TextDark}{\textbf{INGF ($\uparrow$)}} &
\textcolor{TextDark}{\textbf{TTR ($\uparrow$)}} \\
\midrule
\rowcolor{RowMintPurple}
UltraChat & 0.148 & 0.530 & 0.074 \\
ShareGPT  & 0.107 & 0.569 & 0.128 \\
\rowcolor{RowMintPurple}
\dataset (\emph{ours}) & \bfseries 0.056 & \bfseries 0.572 & \bfseries 0.143 \\
\bottomrule
\end{tabular}

\vspace{-10pt}
\end{table}

\paragraph{Human evaluation.}

\begin{table}[htbp]
\centering
\small
\setlength{\tabcolsep}{4pt}
\renewcommand{\arraystretch}{1}
\caption{Human evaluation results. Reported in Align@5 ($\uparrow$), Plaus@5 ($\uparrow$) and AgreeRate\% ($\uparrow$).}
\label{tab:human-eval}
\vspace{-2pt}
\resizebox{0.38\linewidth}{!}{%
\begin{tabular}{l l r}
\toprule
\rowcolor{HeaderMintPurple}
\textcolor{TextDark}{\textbf{Task}} &
\textcolor{TextDark}{\textbf{Metric}} &
\textcolor{TextDark}{\textbf{Score}} \\
\midrule

\rowcolor{RowMintPurple}
Task A & Align@5 ($\uparrow$)           & \bfseries 4.538 $\;(\pm 0.736)$ \\
      & High-Align\% ($\uparrow$)       & \bfseries 91.03\% \\
\midrule

\rowcolor{RowMintPurple}
Task B & Plaus@5 ($\uparrow$)           & \bfseries 4.940 $\;(\pm 0.238)$ \\
      & High-Plaus\% ($\uparrow$)       & \bfseries 100\% \\
\midrule

\rowcolor{RowMintPurple}
Task C & AgreeRate\% ($\uparrow$)       & \bfseries 89.33 \\
\bottomrule
\end{tabular}
}
\end{table}

\begin{table*}[!t]
\centering
\small
\setlength{\tabcolsep}{1pt}
\renewcommand{\arraystretch}{1}

\caption{Personalization score and task completion score by model and synthetic evaluation set. Reported: $\mathrm{mean}_{\mathrm{var}}$.}
\vspace{-3pt}
\label{tab:person_by_model_5datasets}
\resizebox{\textwidth}{!}{%
\begin{tabular}{c ccccc ccccc ccccc}
\toprule
& \multicolumn{5}{c}{\textbf{Llama-3-8b-instruct}} & \multicolumn{5}{c}{\textbf{Qwen3-8b}} & \multicolumn{5}{c}{\textbf{GPT-oss-20b}} \\
\cmidrule(lr){2-6}\cmidrule(lr){7-11}\cmidrule(lr){12-16}

& \texttt{ai2\_arc} & \texttt{IFEval} & \texttt{MBPP} & \texttt{oasst1} & \texttt{ultrachat}
& \texttt{ai2\_arc} & \texttt{IFEval} & \texttt{MBPP} & \texttt{oasst1} & \texttt{ultrachat}
& \texttt{ai2\_arc} & \texttt{IFEval} & \texttt{MBPP} & \texttt{oasst1} & \texttt{ultrachat}\\
\midrule
\multicolumn{16}{c}{\textbf{Personalization Score}} \\ 
\midrule
\rowcolor{RowMintPurple}\textbf{Vanilla}
& $4.62_{0.001}$ & $5.48_{0.013}$ & $5.96_{0.003}$ & $5.30_{0.020}$ & $5.56_{0.003}$
& $4.62_{0.001}$ & $5.48_{0.013}$ & $5.96_{0.003}$ & $5.30_{0.020}$ & $5.56_{0.003}$
& $4.62_{0.001}$ & $5.48_{0.013}$ & $5.96_{0.003}$ & $5.30_{0.020}$ & $5.56_{0.003}$ \\
\textbf{PPOpt.}
& $7.38_{0.135}$ & $6.58_{0.007}$ & $7.90_{0.001}$ & $7.26_{0.001}$ & $7.00_{0.080}$
& $6.60_{0.157}$ & $6.56_{0.013}$ & $7.74_{0.097}$ & $7.00_{0.157}$ & $7.00_{0.205}$
& $7.34_{0.180}$ & $6.52_{0.051}$ & $7.78_{0.007}$ & $6.88_{0.080}$ & $7.26_{0.065}$ \\
\rowcolor{RowMintPurple}\textbf{$\Delta$}
& $2.76_{0.157}$ & $1.10_{0.001}$ & $1.94_{0.007}$ & $1.96_{0.013}$ & $1.44_{0.051}$
& $1.98_{0.180}$ & $1.08_{0.000}$ & $1.78_{0.135}$ & $1.70_{0.289}$ & $1.44_{0.259}$
& $2.72_{0.205}$ & $1.04_{0.115}$ & $1.82_{0.020}$ & $1.58_{0.180}$ & $1.70_{0.097}$ \\
\midrule
\multicolumn{16}{c}{\textbf{Task Completion Score}} \\ 
\midrule
\rowcolor{RowMintPurple}\textbf{Vanilla}
& $9.56_{0.003}$ & $8.28_{0.013}$ & $9.08_{0.013}$ & $7.94_{0.001}$ & $7.88_{0.051}$
& $9.56_{0.003}$ & $8.28_{0.013}$ & $9.08_{0.013}$ & $7.94_{0.001}$ & $7.88_{0.051}$
& $9.56_{0.003}$ & $8.28_{0.013}$ & $9.08_{0.013}$ & $7.94_{0.001}$ & $7.88_{0.051}$ \\
\textbf{PPOpt}
& $9.50_{0.097}$ & $8.20_{0.259}$ & $8.70_{0.180}$ & $7.76_{0.051}$ & $7.70_{0.007}$
& $9.30_{0.065}$ & $7.50_{0.097}$ & $8.84_{0.013}$ & $7.94_{0.097}$ & $7.82_{0.065}$
& $9.48_{0.003}$ & $8.24_{0.320}$ & $8.86_{0.007}$ & $7.80_{0.013}$ & $7.96_{0.051}$ \\
\rowcolor{RowMintPurple}\textbf{|$\Delta$|}
& $0.06_{0.065}$ & $0.08_{0.387}$ & $0.38_{0.097}$ & $0.18_{0.039}$ & $0.18_{0.097}$
& $0.26_{0.097}$ & $0.78_{0.039}$ & $0.24_{0.000}$ & $0.00_{0.115}$ & $0.06_{0.001}$
& $0.08_{0.013}$ & $0.04_{0.461}$ & $0.22_{0.039}$ & $0.14_{0.020}$ & $0.08_{0.205}$ \\
\bottomrule
\end{tabular}}
\end{table*}

We now evaluate our dataset through human judgments across tasks (for all metrics: higher is better). The tasks involve profile--query/feedback alignment (Task A), distractor plausibility (Task B), outcome label fit (Task C); see Appendix~\ref{app:human_eval_appendix}. Align@5 and Plaus@5 report the mean of 5-point Likert ratings for (A) alignment with the task requirements, and (B) plausibility. Further, High-Align\% and High-Plaus\% denote the fraction of instances receiving a high rating ($\ge 4$), respectively. AgreeRate\% is the percentage of instances for which annotators agreed that the output satisfies Task C.  Table~\ref{tab:human-eval} shows strong alignment and plausibility overall. 
These results indicate that the generated responses are consistently perceived as personalized (aligned with user needs) and natural (plausible and coherent). Further details are in the Appendix~\ref{app:human_eval_appendix}.

\begin{table}[htbp]
\centering
\caption{Comparison of personalization and task completion performance across different baselines.}
\label{tab:baseline}
\setlength{\tabcolsep}{3pt}
\scalebox{0.8}{
\begin{tabular}{lccc ccc}
\toprule
\multirow{2}{*}{\textbf{Method}}
& \multicolumn{3}{c}{\textbf{Personalization}}
& \multicolumn{3}{c}{\textbf{Task Completion}} \\
\cmidrule(lr){2-4} \cmidrule(lr){5-7}
& Vanilla & PPOpt & $\Delta$ 
& Vanilla & PPOpt & $|\Delta|$ \\
\midrule
\rowcolor{RowMintPurple}History-augmented prompting
& 5.41 & 6.94 & $1.53_{\,28.28\%}$
& 8.48 & 7.89 & $0.59_{\,6.96\%}$ \\

Persona-based query rewriting \citep{Richardson2023IntegratingSA, Garbacea2025HyPerAlignIP, Zhao2025NextQuillCP}
& 5.41 & 6.22 & $0.81_{\,14.97\%}$
& 8.48 & 7.42 & $1.06_{\,12.50\%}$ \\

\rowcolor{RowMintPurple}Preference-based few-shot ICL \citep{Pitis2024ImprovingCP, Yu2024ICPLFI}
& 5.41 & 6.86 & $1.45_{\,26.80\%}$
& 8.48 & 7.94 & $0.54_{\,6.37\%}$ \\

Controller-guided prompting \citep{Li2024MatryoshkaPL, Wang2024AIPT}
& 5.41 & 6.67 & $1.26_{\,23.29\%}$
& 8.48 & 7.94 & $0.54_{\,6.37\%}$ \\

\midrule
\rowcolor{RowMintPurple}\textbf{PPOpt (\textit{ours})}
& 5.41 & \textbf{7.20} & $\mathbf{1.79_{\,33.09\%}}$
& 8.48 & \textbf{8.26} & $\mathbf{0.22_{\,2.59\%}}$ \\
\bottomrule
\end{tabular}
}
\vspace{-6pt}
\end{table}

\begin{table}[t]
\centering
\small
\setlength{\tabcolsep}{8pt}
\renewcommand{\arraystretch}{1}

\caption{Personalization and task completion performance with different optimizer and deployed model combinations. \texttt{llama-3.3-70b.} denotes \texttt{Llama-3.3-70B-Instruct}, and
\texttt{Claude-Sonnet} denotes \texttt{Claude-Sonnet-4.5}.}
\vspace{-3pt}
\label{tab:optimizer_deployed}

\begin{tabular}{c ccccccc}
\toprule
\multirow{2}{*}{\textbf{Optimizer}} &
\multirow{2}{*}{\textbf{Deployed Model}} &
\multicolumn{3}{c}{\textbf{Personalization}} &
\multicolumn{3}{c}{\textbf{Task Completion}} \\
\cmidrule(lr){3-5}\cmidrule(lr){6-8}
& & \textbf{Vanilla} & \textbf{PPOpt} & $\boldsymbol{\Delta}$ 
  & \textbf{Vanilla} & \textbf{PPOpt} & $\boldsymbol{|\Delta|}$ \\
\midrule

\multirow{5}{*}{\texttt{Llama 3}}
& \texttt{GPT-4o-mini}            & 7.05 & 8.26 &  1.21 & 8.26 & 8.11 & 0.15 \\
& \texttt{GPT-5.2}                & 7.66 & 8.66 &  1.00 & 8.53 & 8.24 & 0.29 \\
& \texttt{llama-3.3-70b.} & 6.78 & 8.08 &  1.30 & 8.11 & 7.96 & 0.15 \\
& \texttt{Qwen3-32B}              & 7.50 & 8.56 &  1.06 & 7.98 & 8.05 &  0.07 \\
& \texttt{Claude-Sonnet}      & 7.75 & 8.61 &  0.86 & 8.36 & 8.34 & 0.02 \\
\midrule

\multirow{5}{*}{\texttt{Qwen3}}
& \texttt{GPT-4o-mini}            & 6.93 & 8.11 &  1.18 & 8.21 & 8.27 &  0.06 \\
& \texttt{GPT-5.2}                & 7.82 & 8.53 &  0.71 & 8.43 & 8.06 & 0.37 \\
& \texttt{llama-3.3-70b.} & 6.98 & 7.89 &  0.91 & 8.06 & 7.76 & 0.30 \\
& \texttt{Qwen3-32B}              & 7.45 & 8.34 &  0.89 & 7.75 & 7.99 &  0.24 \\
& \texttt{Claude-Sonnet}      & 7.70 & 8.42 &  0.72 & 8.57 & 8.47 & 0.10 \\
\midrule

\multirow{5}{*}{\texttt{GPT-oss}}
& \texttt{GPT-4o-mini}            & 6.93 & 7.42 &  0.49 & 8.28 & 8.13 & 0.15 \\
& \texttt{GPT-5.2}                & 7.71 & 8.14 &  0.43 & 8.39 & 8.10 & 0.29 \\
& \texttt{llama-3.3-70b.} & 6.68 & 7.38 &  0.70 & 7.97 & 7.97 &  0.00 \\
& \texttt{Qwen3-32B}              & 7.52 & 7.98 &  0.46 & 7.91 & 7.93 &  0.02 \\
& \texttt{Claude-Sonnet}      & 7.74 & 8.14 &  0.40 & 8.52 & 8.50 & 0.02 \\

\bottomrule

\end{tabular}

\end{table}

\begin{table}[t]
\centering
\small
\setlength{\tabcolsep}{6pt}
\renewcommand{\arraystretch}{1}

\caption{Personalization and task completion scores on the real-world test set.}
\vspace{-3pt}
\label{tab:optimizer_models_person_tc}

\scalebox{0.9}{
\begin{tabular}{lccc}
\toprule
\textbf{Method} & \textbf{Llama-3-8b-instruct} & \textbf{Qwen3-8b} & \textbf{GPT-oss-20b} \\
\midrule

\multicolumn{4}{c}{\textbf{Personalization}} \\
\midrule
\rowcolor{RowMintPurple}\textbf{Vanilla}        & 5.49 & 5.49 & 5.49 \\
\textbf{PPOpt}      & 7.20 & 7.10 & 7.35 \\
\rowcolor{RowMintPurple}\textbf{$\Delta$}   & 1.71 & 1.61 & 1.86 \\
\midrule

\multicolumn{4}{c}{\textbf{Task Completion}} \\
\midrule
\rowcolor{RowMintPurple}\textbf{Vanilla}        & 8.09 & 8.09 & 8.09 \\
\textbf{PPOpt}      & 8.05 & 8.08 & 7.99 \\
\rowcolor{RowMintPurple}\textbf{$|\Delta|$}   & 0.04 & 0.01 & 0.10 \\
\bottomrule
\end{tabular}
}

\vspace{-5pt}
\end{table}

\vspace{-3pt}
\subsection{Experimental Setup}
\label{subsec:exp_setup}

\paragraph{Models \& Baselines.} We use a broad set of open- and closed-source, instruction-tuned LLMs for evaluation. Detailed model lists are provided in the Appendix~\ref{app:eval_model}.
We compare our method with the vanilla prompt and the following representative personalization approaches: (1)~history-augmented prompting (Simple Concat),  persona-based query rewriting~\citep{Richardson2023IntegratingSA, Garbacea2025HyPerAlignIP, Zhao2025NextQuillCP}, (3)~preference-based few-shot in-context learning~\citep{Pitis2024ImprovingCP, Yu2024ICPLFI}, and (4)~controller-guided Prompting~\citep{Li2024MatryoshkaPL, Wang2024AIPT}. More details are provided in in Appendix~\ref{app:baseline_implementation}).

\paragraph{Datasets}

We evaluate our method in two ways: (1)~synthetic personalized interaction datasets and (2)~a real-world test set. 
The synthetic benchmarks consist of multi-turn trajectories constructed from multiple public datasets (i.e., \texttt{ai2\_arc}, \texttt{IFEval}, \texttt{oasst1}, \texttt{ultrachat}), with additional held-out datasets used only for evaluation. 
The real-world test set contains high-fidelity personalized interactions processed into $(\mathcal{H}_u, \tilde{q}^{\mathrm{init}}_u)$ pairs. 
Implementation and dataset details are in Appendix~\ref{app:dataset}.

\paragraph{Evaluation Metrics} We evaluate model performance using two metrics by LLM-as-a-Judge \citep{zheng2023judging}: a \emph{personalization score} and a \emph{task completion score}. (1) \textit{Personalization score:}
To measure personalization quality, we provide the user profile together with the model output to an LLM-based judge, which assigns a score indicating how well the response aligns with the specified user preferences. (2) \textit{Task completion score:}
To measure task performance, we provide ground-truth answers to an LLM-based judge, which scores the model output based on task correctness and completeness.  We report the mean and variance over 3 runs.

\begin{figure}[htbp]
    \centering
    \includegraphics[width=0.7\linewidth]{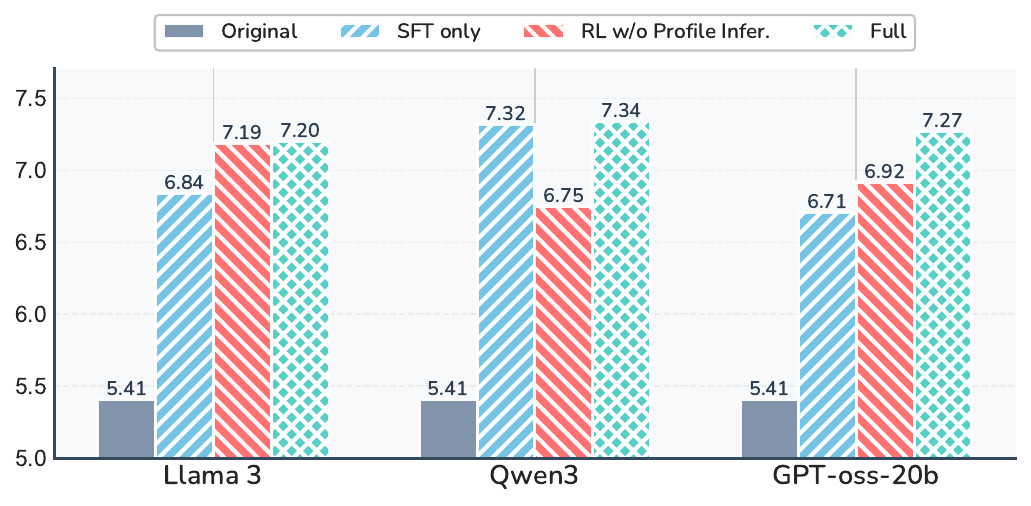}
    \vspace{10pt}
    \caption{Ablation study of \methodshort. We report personalization scores under different training settings: SFT only, RL w/o the profile inference reward, and our full setting.}
    \label{fig:ablation}
\end{figure}

\subsection{Experiment Results}
\label{subsec:baseline}

\paragraph{Overall Performance.} We report the overall personalization quality and task completion in Table~\ref{tab:person_by_model_5datasets} (synthetic evaluation) and Table~\ref{tab:optimizer_models_person_tc} (real-world evaluation). The results are based on the deployed model of \texttt{Llama-3.3-70b-instruct} (selected because it is an open-source model). The results show that \methodshort\  with \textit{all three optimizers} achieves substantial improvements in personalization. At the same time, we find that the differences in task completion are very small across methods, indicating that the impact of \methodshort\ on task completion is acceptable.

\paragraph{Baseline comparison.} Table~\ref{tab:baseline} compares \methodshort against several representative prompting and rewriting baselines in terms of personalization quality and task completion. All baselines improve personalization over the vanilla prompts (e.g., $+0.81$ to $+1.53$), but these gains typically come with a non-trivial drop in task completion ($|\Delta|=0.54$--$1.06$), indicating a trade-off where stronger personalization steering can lightly hurt task completion quality. In contrast, \methodshort achieves the largest gain personalization ($\Delta=1.79$, $33.09 \%$) while preserving task completion most effectively ($|\Delta|=0.22$, $2.59 \%$), thereby substantially reducing the performance degradation observed in prior methods. This suggests \methodshort better captures user-specific constraints without over-constraining the underlying model, thus yielding improved personalization with minimal impact on task success.

\paragraph{Robustness across LLM Backbones.} From Table~\ref{tab:optimizer_deployed}, we observe that \methodshort\ consistently improves personalization across all optimizer–model combinations. Notably, optimization can elevate the personalization performance of weaker deployed models to a level comparable with that of stronger models, despite clear differences in their original personalization capabilities. Meanwhile, the impact on task completion remains minimal, with only small changes observed. 

\paragraph{Ablation Study.} Figure~\ref{fig:ablation} shows that SFT alone already provides a strong boost in personalization across all optimizers, establishing an effective cold-start prior. However, RL without the profile inference reward leads to inconsistent gains, and even degrades performance for some models (e.g., Qwen3), probably indicating unstable optimization without explicit guidance. In contrast, the full setting consistently achieves the best and most stable performance, demonstrating that the profile inference reward plays a crucial role in stabilizing RL and further improving personalization.

\subsection*{Acknowledgments}
This work has been supported by the German Federal Ministry of Education and Research (Grant: 01IS24082).

\bibliographystyle{ACM-Reference-Format}
\bibliography{references}

\newpage
\appendix
\onecolumn

\section*{Generative AI Usage}
We used generative AI tools during the research and writing process in a limited, assistive manner. Specifically, we used such tools (i) to polish and proofread text (e.g., improving clarity, grammar, and conciseness), (ii) to help debug and refine code implementations (e.g., diagnosing errors and suggesting fixes), and (iii) to refine and polish figures created by the authors (e.g., improving layout, labeling, and visual consistency), without generating figures from scratch. All scientific decisions, experimental design, data generation choices, result interpretation, and final manuscript content were reviewed and approved by the authors.

\section*{Ethics Statement}
The construction of our dataset follows best-practice for ethical research \cite{rivers2014ethical}. The dataset construction and usage were approved as ethically unproblematic by the corresponding ethics commission (i.e., institutional review board) of LMU Munich
(approval number: ETH-SOM-048). Further, we respect the privacy of users and only report aggregate results throughout our paper. Although we believe the intended use of this work is largely positive, there exists potential
for misuse.

\newpage
\section{Algorithm}
\label{app:alg}

\begin{algorithm}[htbp]
\caption{Synthetic Personalized Interaction Generation}
\label{alg:synth}
\begin{algorithmic}[1]
\Require Persona bank $\mathcal{P}$, task datasets $\mathcal{D}$, masking policy $\pi_{\mathrm{mask}}$, stylization prob. $\rho$, max turns $T$
\Require Preference spec compiler $\Psi$, models $M_{\mathrm{user}}, M_{\mathrm{asst}}, M_{\mathrm{dist}}$, noise strengths $\alpha_q,\alpha_f$
\Require Noise application probabilities $p_q, p_f$

\Statex \textcolor{Purple!75}{\footnotesize\bfseries $\triangleright$ Persona sampling \& partial observability (persona $\rightarrow$ observed features)}
\State Sample persona $p \sim \mathcal{P}$
\State Sample observed features $o \sim \mathrm{Sample}(p; \pi_{\mathrm{mask}})$ \Comment{Eq.~\eqref{eq:obs-features}}

\Statex \textcolor{Purple!75}{\footnotesize\bfseries $\triangleright$ Preference compilation (observed features $\rightarrow$ system-level spec)}
\State Compile preference spec $s \gets \mathrm{Compile}(o; \Psi)$ \Comment{Eq.~\eqref{eq:compile}}

\Statex \textcolor{Purple!75}{\footnotesize\bfseries $\triangleright$ Seed query sampling \& optional persona stylization}
\State Sample seed query $q \sim \mathcal{D}$
\State $\tilde{q} \gets \mathrm{Stylize}(q,s;M_{\mathrm{user}})$ with prob. $\rho$; otherwise $\tilde{q}\gets q$ \Comment{Eq.~\eqref{eq:stylize}}

\Statex \textcolor{Purple!75}{\footnotesize\bfseries $\triangleright$ Query perturbation (distractor injects realistic noise)}
\State with prob. $p_q$, $\tilde{q} \gets \mathrm{Noisify}(\tilde{q};M_{\mathrm{dist}},\alpha_q)$ \Comment{Eq.~(17)}

\State Initialize history $h_0 \gets \emptyset$

\Statex \textcolor{Purple!75}{\footnotesize\bfseries $\triangleright$ Trajectory synthesis (query $\rightarrow$ response $\rightarrow$ feedback $\rightarrow$ follow-up/stop)}
\For{$t=1$ to $T$}
    \State Set current prompt $\tilde{q}_t \gets \tilde{q}$
    \State Assistant response $a_t \gets M_{\mathrm{asst}}(\tilde{q}_t, h_{t-1})$ \Comment{Eq.~(11)}
    \State User feedback $f_t \gets M_{\mathrm{user}}(s,\tilde{q}_t,a_t,h_{t-1})$ \Comment{Eq.~(12)}

    \Statex \textcolor{Purple!75}{\footnotesize\bfseries $\triangleright$ Feedback perturbation (optional noisy follow-up signals)}
    \State with prob. $p_f$, $f_t \gets \mathrm{Noisify}(f_t;M_{\mathrm{dist}},\alpha_f)$ \Comment{Eq.~(18)}

    \State Determine follow-up $(\mathrm{stop}, \tilde{q}_{t+1}) \gets \mathrm{Decide}(f_t)$

    \Statex \textcolor{Purple!75}{\footnotesize\bfseries $\triangleright$ Outcome-derived supervision (stop $\Rightarrow$ positive)}
    \State Outcome label $y_t \gets \mathbb{I}[\mathrm{stop}]$ \Comment{Eq.~\eqref{eq:label}}
    \State Save instance $(o, s, h_{t-1}, \tilde{q}_t, a_t, f_t, y_t)$

    \If{$\mathrm{stop}$}
        \State \textbf{break}
    \EndIf
    \State Update history $h_t \gets h_{t-1}\cup\{(\tilde{q}_t,a_t,f_t)\}$
    \State Update prompt $\tilde{q} \gets \tilde{q}_{t+1}$
\EndFor
\State \Return synthesized instances
\end{algorithmic}
\end{algorithm}

\clearpage
\section{Method details}
\label{app:method}

\begin{figure}[h]
    \centering    \includegraphics[width=0.7\linewidth]{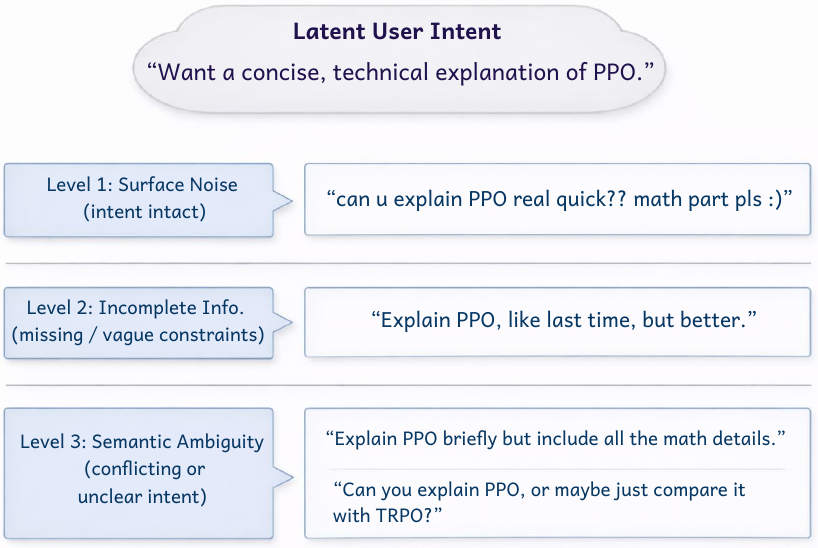}
    \vspace{15pt}
    \caption{Examples of distractor perturbations at three levels. All examples share the same latent user intent, while the observed queries exhibit increasing degrees of syntactic noise, missing execution-relevant constraints, or semantic ambiguity.}
    \label{fig:distractor}
\end{figure}

\vspace{10pt}

\begin{figure}[h]
    \centering
    \includegraphics[width=0.8\linewidth]{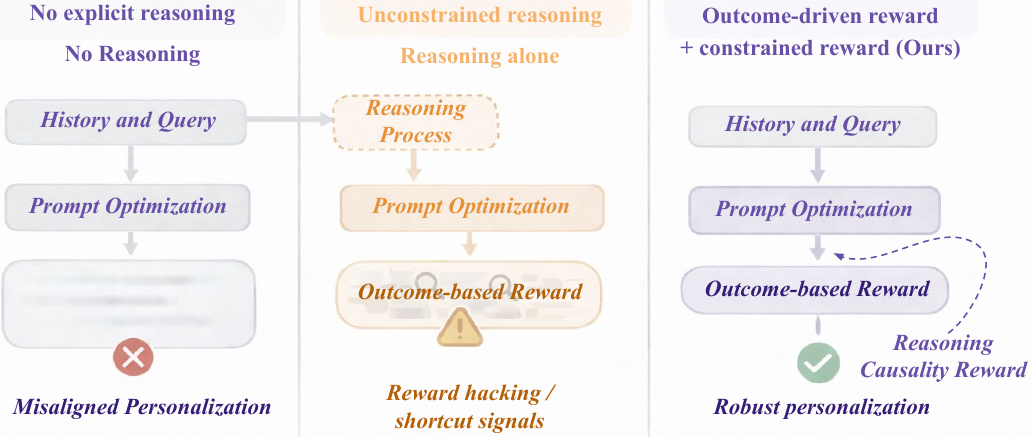}
    \vspace{15pt}
    \caption{Reasoning and reward design for personalized prompt optimization. Unconstrained or absent reasoning leads to misalignment or reward hacking, whereas combining outcome-driven rewards with constrained reasoning causality yields robust personalization.}
    \label{fig:rl_training}
\end{figure}

\clearpage
\section{Implementation details}
\label{app:implementation_details}

\paragraph{Filtering low-quality pseudo targets for SFT.}
Since $(z^\star_u, q^\star_u)$ are synthesized and may be noisy, we apply lightweight filtering to remove non-effective or degenerate rewrites before SFT. Concretely, we discard samples where the rewritten prompt $q^\star_u$ is near-identical to $\tilde{q}^{\mathrm{init}}_u$ (e.g., trivial paraphrases or copies), overly generic (e.g., adding boilerplate without introducing user-specific constraints), or leads to no measurable improvement under an automatic preference-aware comparison (e.g., $\mathcal{J}_{\mathrm{task}}$ fails to prefer the response induced by $q^\star_u$ over that induced by $\tilde{q}^{\mathrm{init}}_u$).
This filtering step improves the signal-to-noise ratio of the behavioral prior learned by SFT and reduces the tendency to imitate superficial prompt edits.

\paragraph{Implementation details of baselines}
\label{app:baseline_implementation} 

\begin{itemize}[leftmargin=1.5em]
\item \textbf{History-augmented prompting (Concat).}  
The user’s conversation history is directly concatenated with the current query and provided to the assistant model for response generation. The assistant’s output is subsequently evaluated by a judge model to obtain the final score.

\item \textbf{Persona-based query rewriting} \citep{Richardson2023IntegratingSA, Garbacea2025HyPerAlignIP, Zhao2025NextQuillCP}.  
A high-level user persona is first induced from the conversation history using a helper model (\texttt{llama-3.1-8b-instruct}). The inferred persona is then concatenated with the current query, and the combined input is passed to the assistant model for response generation. The resulting output is scored by the judge model.

\item \textbf{Preference-based few-shot in-context learning} \citep{Pitis2024ImprovingCP, Yu2024ICPLFI}.  
A helper model (\texttt{GPT-4o-mini}) constructs a small set of preference-oriented few-shot examples (e.g., chosen vs.\ rejected responses) from the user’s interaction history. The current query, together with the generated preference pairs, is prepended as in-context demonstrations to guide personalized generation by the assistant model.

\item \textbf{Controller-guided prompting} \citep{Li2024MatryoshkaPL, Wang2024AIPT}.  
The conversation history and current query are provided to an auxiliary controller model (\texttt{GPT-4o-mini}), which generates context-enhanced personalized guidance by rewriting the query based on the interaction history. The rewritten query is then passed to the assistant model for response generation.
\end{itemize}
Unless otherwise specified, the assistant model is \texttt{llama-3.3-70b-instruct}, and the judge model is \texttt{GPT-5.2}.

\paragraph{Training Details}

We adopt a two-stage training paradigm for \methodshort. 
In the first stage, we perform Supervised Fine-Tuning (SFT) with Low-Rank Adaptation (LoRA) \citep{hu2022lora} using rank $r=32$. 
The model is trained to generate structured outputs that include user preference reasoning and optimized prompts. 
During SFT, we apply cross-entropy loss over completion tokens only. 
A cosine learning rate schedule is used, with a peak learning rate of $2 \times 10^{-6}$, a minimum learning rate of $5 \times 10^{-7}$, and $50$ warmup steps.

In the second stage, we apply Group Relative Policy Optimization (GRPO) for reinforcement learning.
For each training state, we sample $K=8$ completions from the current policy using temperature $\tau = 0.7$ and nucleus sampling with $p=0.9$.
We employ \texttt{GPT-4o-mini} as the judge model to compute rewards, which evaluate both the accuracy of inferred user profiles against ground-truth personas and the quality improvement of optimized queries via pairwise comparison with the original queries.

\section{Case Study}
\label{app:case_study}

\newtcbox{\stmt}{on line,
  boxsep=0pt,
  left=1pt, right=1pt, top=1pt, bottom=1pt,
  colback=yellow!22,
  colframe=yellow!22,
  arc=1pt,
  boxrule=0pt
}

\newtcbox{\infer}{on line,
  boxsep=0pt,
  left=1pt, right=1pt, top=1pt, bottom=1pt,
  colback=cyan!16,
  colframe=cyan!16,
  arc=1pt,
  boxrule=0pt
}

\newtcbox{\impprompt}{on line,
  boxsep=0pt,
  left=1pt, right=1pt, top=1pt, bottom=1pt,
  colback=green!17,
  colframe=green!17,
  arc=1pt,
  boxrule=0pt
}

\begin{tcolorbox}[
  colback=gray!3,
  colframe=gray!60,
  boxrule=0.6pt,
  arc=2pt,
  left=6pt,right=6pt,top=6pt,bottom=6pt,
  breakable,
  enhanced jigsaw
]
\small
\textbf{SFT Synthesized Reasoning-Chain}\\
\textcolor{gray!70}{\textbf{Legend:} \stmt{Statement} \quad \infer{Derived conclusion} \quad \impprompt{Improved prompt}}

\medskip
<REASONING>Across the threads, the user consistently signals \infer{urgency and a preference for a single best, definitive answer}. They repeatedly say \stmt{“I’m in a rush,”} ask for \stmt{“single best pick,”} and want the \stmt{“fastest definitive” check}, which supports \infer{time pressure, concise output, and low tolerance for ambiguity}. They are \infer{risk-averse and safety-focused}: in the school deviled-egg question they prioritize \stmt{“lowest-allergen… won’t introduce… choking risks,”} and in the team-building plan they explicitly say \stmt{“I’m risk-averse”} and request a \stmt{“safety/risk checklist + inclusivity notes,”} showing a \infer{strong risk posture and a QA-like mindset (checklists, edge cases, verification steps)}. They also prefer \infer{precision and concrete verification} (\stmt{“what options should I see… exact device name…} \stmt{basically a hard no?”}), and they push back on \stmt{overbroad claims} (\stmt{“you said ‘no allergens,’ but peppers can still be an issue”}), indicating \infer{zero-tolerance for inaccuracies}. The user’s context often involves \stmt{education/classroom constraints} (\stmt{“school setting,” “classroom-safe”}), and they want outputs that are \infer{immediately usable} (\stmt{printable run-of-show, one page, mobile-friendly}). Tone is \infer{direct but not hostile—more pragmatic and warm}—seeking \infer{minimal context and actionable guidance} \infer{rather than long explanations}.</REASONING><PROMPT>\impprompt{I’m on a deadline--give me a very concise,} \impprompt{risk-averse summary of the main legal risks for authors when (a)} \impprompt{writing book reviews and (b) being} \impprompt{involved with/mentioned alongside Stonewall} \impprompt{Press. Bullet points only: defamation/libel,} \impprompt{copyright/fair use, trademark, privacy/publicity, and any contract/endorsement/conflict-of-interest issues.} \impprompt{Then give a single “safest default” checklist (do/don’t) I can follow. If you’re not sure} \impprompt{about Stonewall Press specifics,} \impprompt{say so and tell me the fastest facts I should verify.}</PROMPT>
\end{tcolorbox}

\section{Evaluation Dataset}
\label{app:dataset}

We use two sets of datasets for the experiments: \textbf{(1) Synthetic personalized interaction benchmark.}
We construct a synthetic multi-turn trajectory benchmark as the test split for evaluation, covering queries from five datasets. The three datasets in Table~\ref{tab:dataset_query_domain} are \texttt{ai2\_arc} \citep{allenai:arc}, \texttt{MBPP} \citep{austin2021program}, and \texttt{ultrachat\_200k} \citep{ding2023enhancing}. 
We additionally include \texttt{IFEval} \citep{zeng2024llmbar} and \texttt{oasst1} \citep{kopf2023openassistant} as held-out evaluation datasets, to reduce potential distributional advantages associated with dataset reuse during training. \textbf{(2) Real-world test set.}
In addition to synthetic benchmarks, we evaluate our method on a high-fidelity dataset that closely simulates real-world personalized interactions. We preprocess the resulting interactions into $(\mathcal{H}_u, \tilde{q}^{\mathrm{init}}_u)$ pairs following the same protocol used throughout our experiments.

\paragraph{Real-world dataset construction.}
Instead of directly relying on raw user logs, we construct this dataset through controlled human-in-the-loop collection to preserve realism while ensuring diversity and reproducibility.

We recruit a single cohort of human experts via Prolific\footnote{\url{https://www.prolific.com/}}, all of whom are experienced in interacting with LLMs and collectively cover a broad range of domains. Participants are compensated at a rate of at least \$10.89 per hour (above minimum wage), and 99 human experts contribute to the data editing process. Participants gave informed consent. The data collection was reviewed by the institutional review board (ethics commission) at LMU Munich
(approval number: ETH-SOM-048).

 We first generate interaction trajectories using our automated generation pipeline. For ease of inspection and editing, each trajectory is formatted as a structured \texttt{Markdown} document. The documents are then distributed to participants through Qualtrics\footnote{\url{https://www.qualtrics.com/}} as survey tasks. Human experts are instructed to edit and refine the user-side content only, including the initial queries and subsequent user feedback, according to specified user profiles. After editing, participants upload the revised Markdown files back to the survey, enabling complete and consistent data collection. This process improves realism and behavioral fidelity while preserving the original task intent and interaction structure.

\section{Evaluation Model}
\label{app:eval_model}

We use a broad set of instruction-tuned LLMs as optimizers, including Llama-3-8B-instruct~\cite{llama3modelcard}, Qwen3-8B~\cite{qwen3technicalreport}, and GPT-oss-20B~\cite{openai2025gptoss120bgptoss20bmodel}. We evaluate deployment-time performance on GPT-4o-mini~\cite{openai2024hello-gpt4o}, GPT-5.2~\cite{openai2025introducing-gpt5-2}, Llama-3.3-70B-instruct~\cite{llama3modelcard}, Qwen3-32B~\cite{qwen3technicalreport}, and Claude-Sonnet-4.5~\cite{anthropic2025claude-sonnet-4-5}.

\section{Details of \dataGeneration}
\label{app:details_data_generation}

\paragraph{Open-ended values via feature expansion.} To avoid an artificially closed-world assumption over persona attributes, we augment each categorical feature dimension with a special \texttt{others} option. This serves as an open-vocabulary placeholder. When a dimension samples \texttt{others}, we dynamically invoke an LLM to synthesize a novel category value that is not present in the predefined taxonomy. We then treat it as an instantiated free-form category for that persona and trajectory.  This mechanism yields an \emph{open-vocabulary} extension of categorical attributes, enabling \emph{unbounded} expansion of the persona space.

\paragraph{Three categories of persona feature dimension.} (i) \textit{Basic user profile attributes}, capturing relatively stable user characteristics (e.g., role, seniority, education level, and domain expertise).
(ii) \textit{Interaction and behavioral traits}, describing how users interact with the system and express intent (e.g., risk posture, time sensitivity, learning style, and communication preferences).
(iii) \textit{output/response constraints}, specifying requirements on the system’s responses (e.g., response length, formatting structure, explanation depth, and citation preference). 

\paragraph{Illustrative example of user query transformation.}
For the same underlying task, a user may sometimes issue a clean request:
\begin{quote}
\small \textbf{Seed/clean:} \texttt{``Summarize this document in bullet points.''}
\end{quote}
Sometimes, a user may also use a prompt that is idiosyncratic (e.g., using abbreviated style, missing punctuation, etc.), which is reflected in the stylization:
\begin{quote}
\small \textbf{Stylized:} \texttt{``Quick summary in bullet points please keep it short and practical, like last time.''}
\end{quote}
By sampling stylization with probability $\rho$, our synthetic trajectories naturally contain both modes, yielding more realistic histories where preferences must be inferred from \emph{occasional} signals rather than being always explicit.

\begin{figure}[h]
    \centering
    \includegraphics[width=1\linewidth]{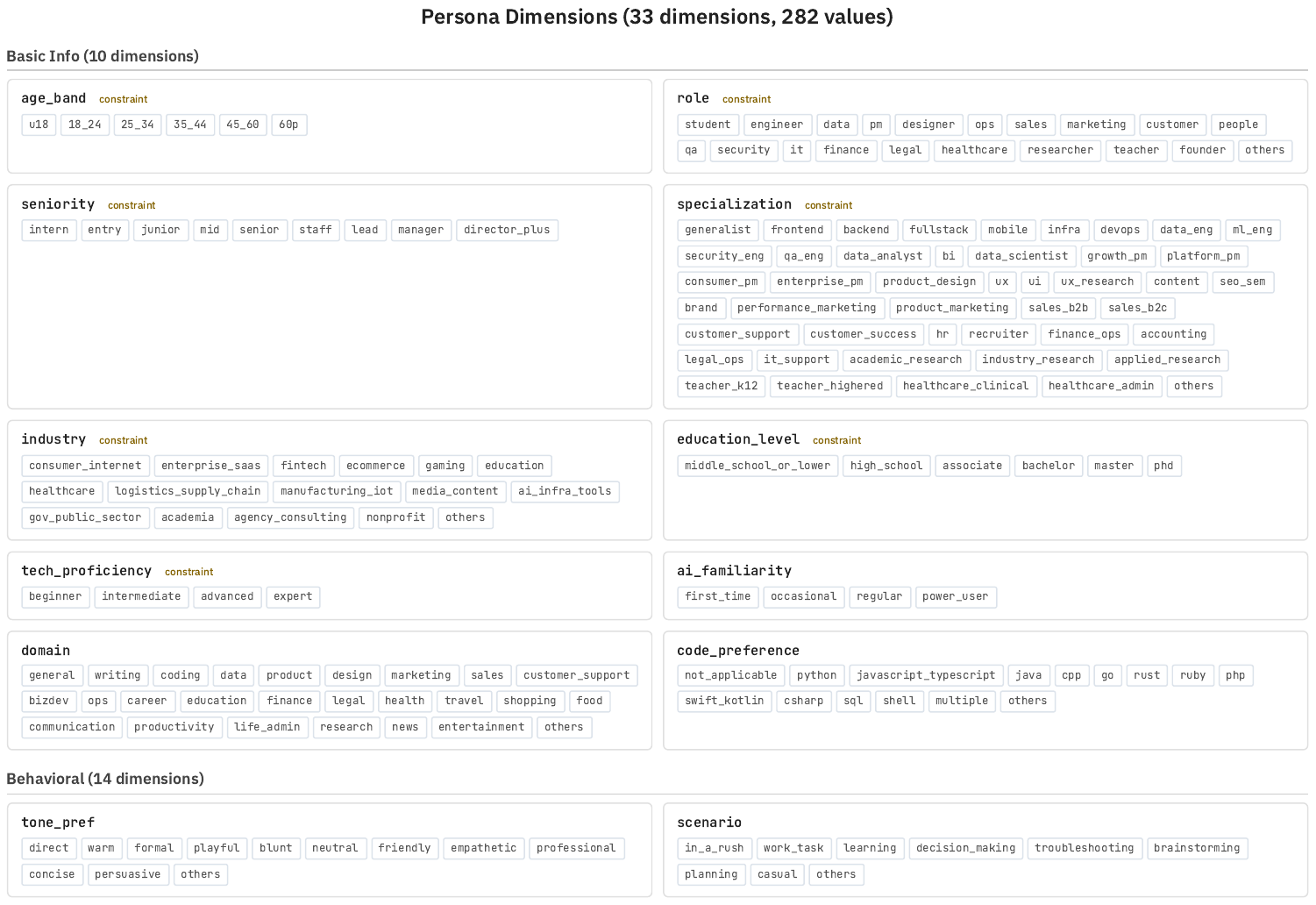}
    \caption{Dimensions and values covered by \dataGeneration (1).}
    \label{fig:dimension_value_1}
\end{figure}

\begin{figure}[h]
    \centering
    \includegraphics[width=1\linewidth]{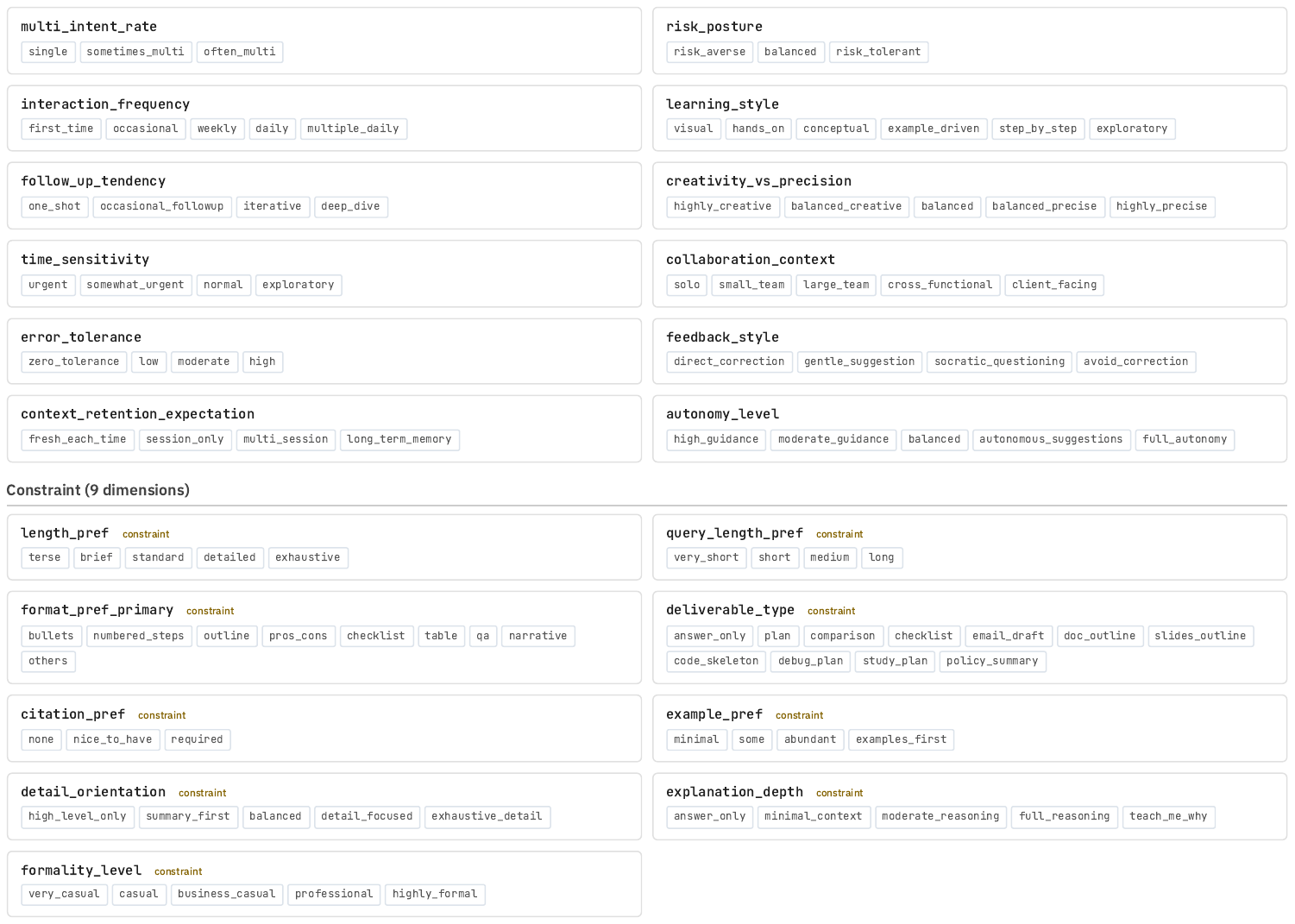}
    \caption{Dimensions and values covered by \dataGeneration (2).}
    \label{fig:dimension_value_2}
\end{figure}

\newpage

\clearpage

\section{\dataset Details}
\label{app:personaatlas}

\begin{figure}[h]
    \centering
    \includegraphics[width=0.9\linewidth]{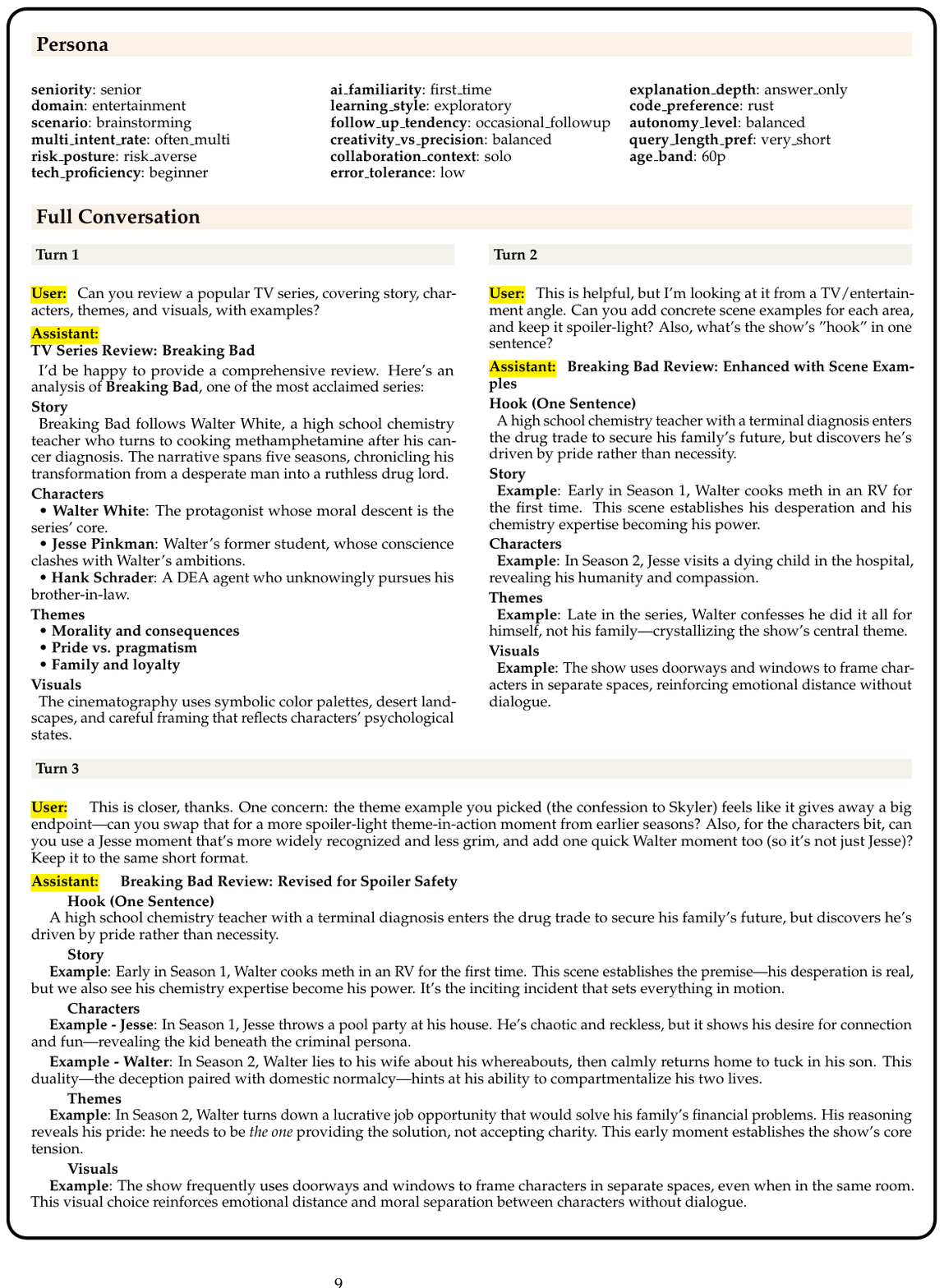}
    \vspace{15pt}
    \caption{\dataset case by \dataGeneration (1).}
    \label{fig:PersonaAtlas_1}
\end{figure}

\begin{figure}[h]
    \centering
    \includegraphics[width=0.9\linewidth]{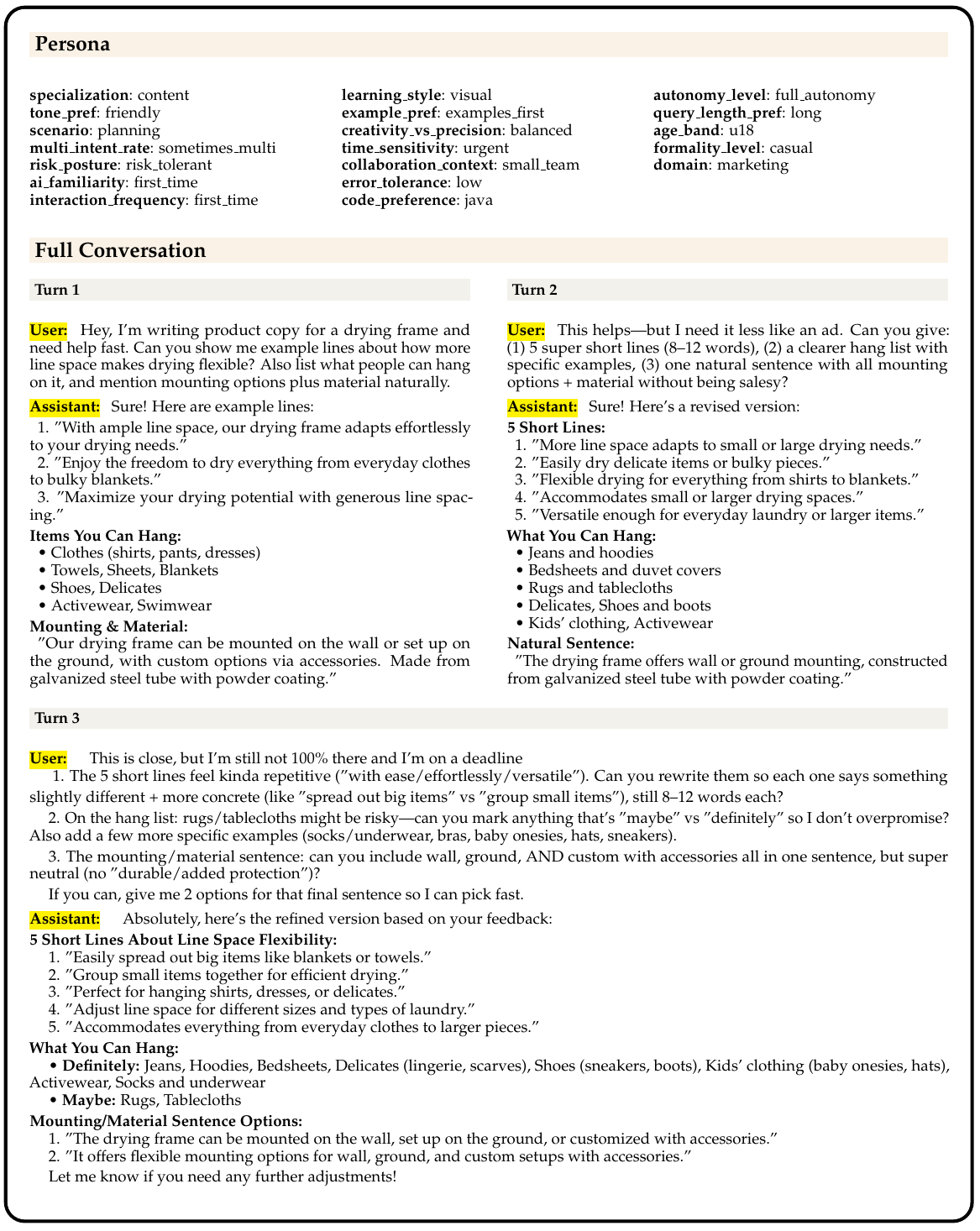}
     \vspace{15pt}
    \caption{\dataset case by \dataGeneration (2).}
    \label{fig:PersonaAtlas_2}
\end{figure}

\begin{figure}[h]
    \centering
    \includegraphics[width=0.9\linewidth]{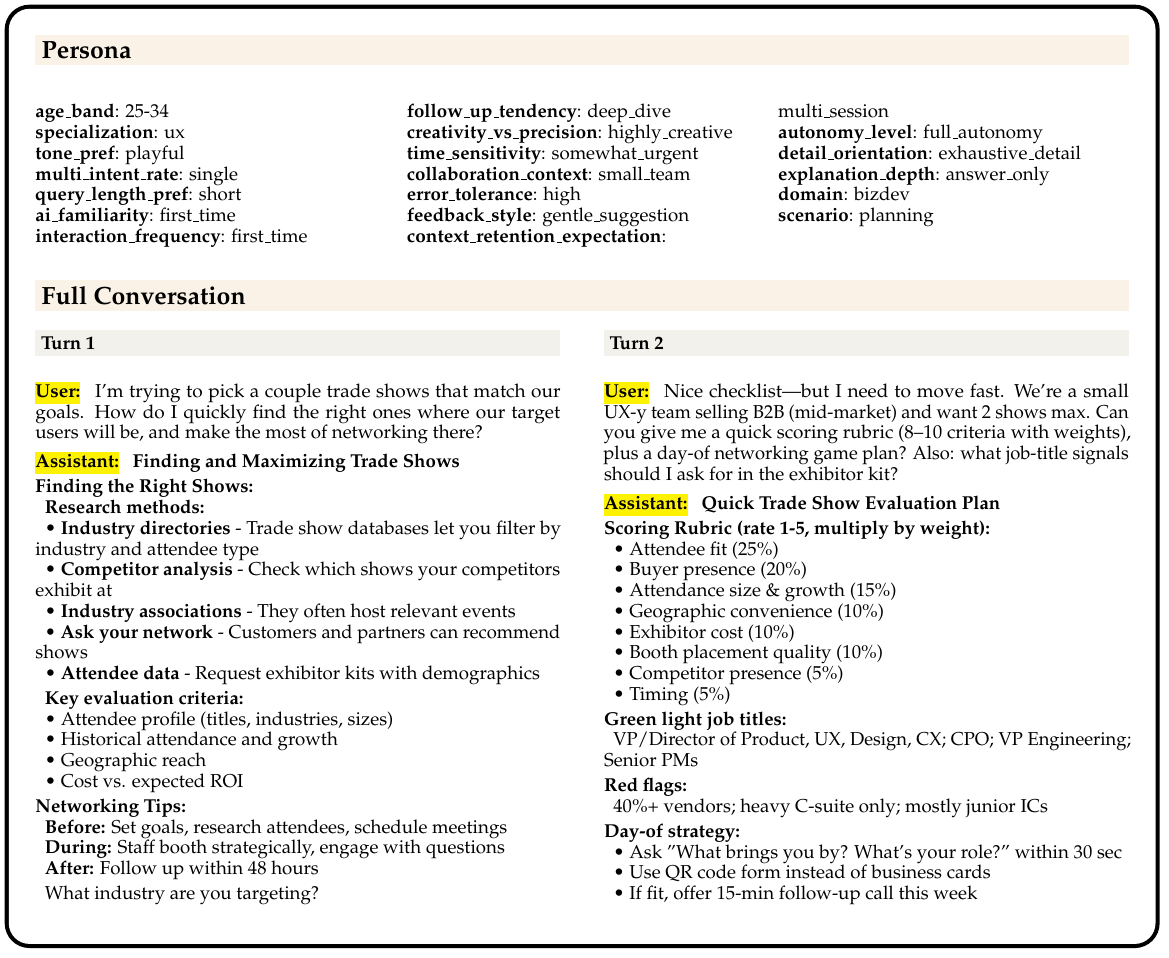}
     \vspace{15pt}
    \caption{\dataset case by \dataGeneration (3).}
    \label{fig:PersonaAtlas_3}
\end{figure}

\begin{figure}[h]
    \centering
    \includegraphics[width=0.9\linewidth]{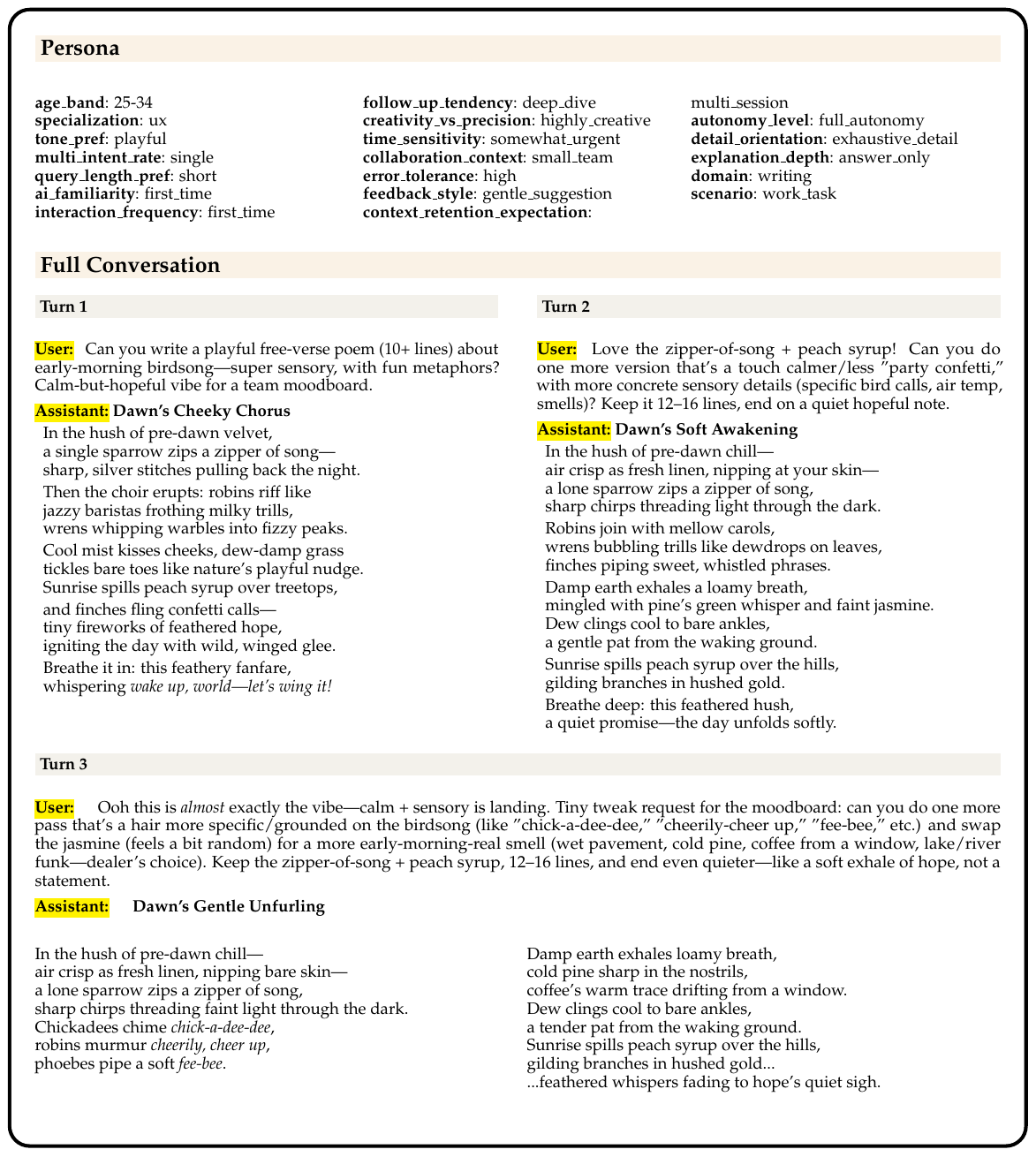}
     \vspace{15pt}
    \caption{\dataset case by \dataGeneration (4).}
    \label{fig:PersonaAtlas_4}
\end{figure}

In this section, we present sample user–assistant interaction data generated in batch using a unified generation rule set.
Each instance is generated based on a predefined user persona to ensure consistency in tone, content strategy, and response style, as shown in Figure~\ref{fig:PersonaAtlas_1}, \ref{fig:PersonaAtlas_2},
\ref{fig:PersonaAtlas_3}, and \ref{fig:PersonaAtlas_4}.

\section{Details of Human Evaluation}
\label{app:human_eval_appendix}

\begin{figure}[h]
    \centering
    \includegraphics[width=1\linewidth]{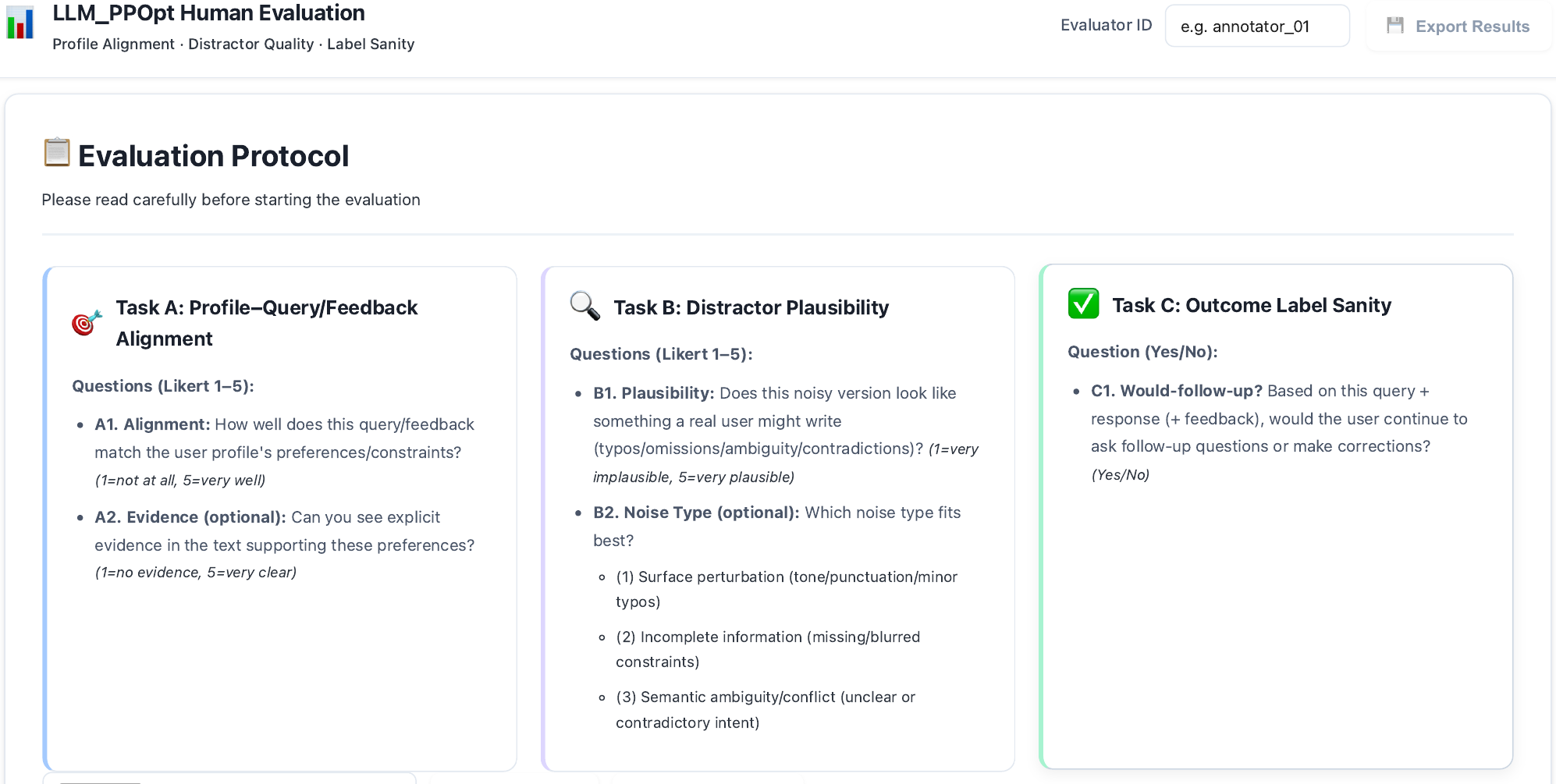}
     \vspace{15pt}
    \caption{Human evaluation protocol for the quality analysis of synthetic data via \dataGeneration.}
    \label{fig:human_protocal}
\end{figure}

We conduct a human evaluation to assess the quality of PersonaGym synthetic data with the protocol (Figure~\ref{fig:human_protocal}). For each sampled instance, annotators are shown the user profile and the corresponding query/feedback, along with the distractor perturbations type (when applicable) and the associated response context. Figure~\ref{fig:human_protocal} summarizes the exact questions and scales used.

\paragraph{Task A: Profile--Query/Feedback Alignment.}
Annotators rate how well the query/feedback matches the displayed user profile's preferences and constraints (A1, Likert 1--5; 1 = not at all, 5 = very well). They rate how much explicit evidence in the text supports those preferences (A2, Likert 1--5; 1 = no evidence, 5 = very clear).

\paragraph{Task B: Distractor Plausibility.}
When a noisy version is provided, annotators rate whether it looks like something a real user might write (B1, Likert 1--5; 1 = very implausible, 5 = very plausible). They optionally label the distractor perturbations type (B2): (1) surface perturbation (tone/punctuation/minor typos), (2) incomplete information (missing/blurred constraints), or (3) semantic ambiguity/conflict (unclear or contradictory intent).

\paragraph{Task C: Outcome Label Fit.}
Annotators answer a binary question on whether the user would likely continue with follow-up questions or corrections given the query and response context (C1, Yes/No), which is used to validate the outcome labels used in our synthetic interaction data.

\begin{figure}[h]
    \centering
    \includegraphics[width=0.7\linewidth]{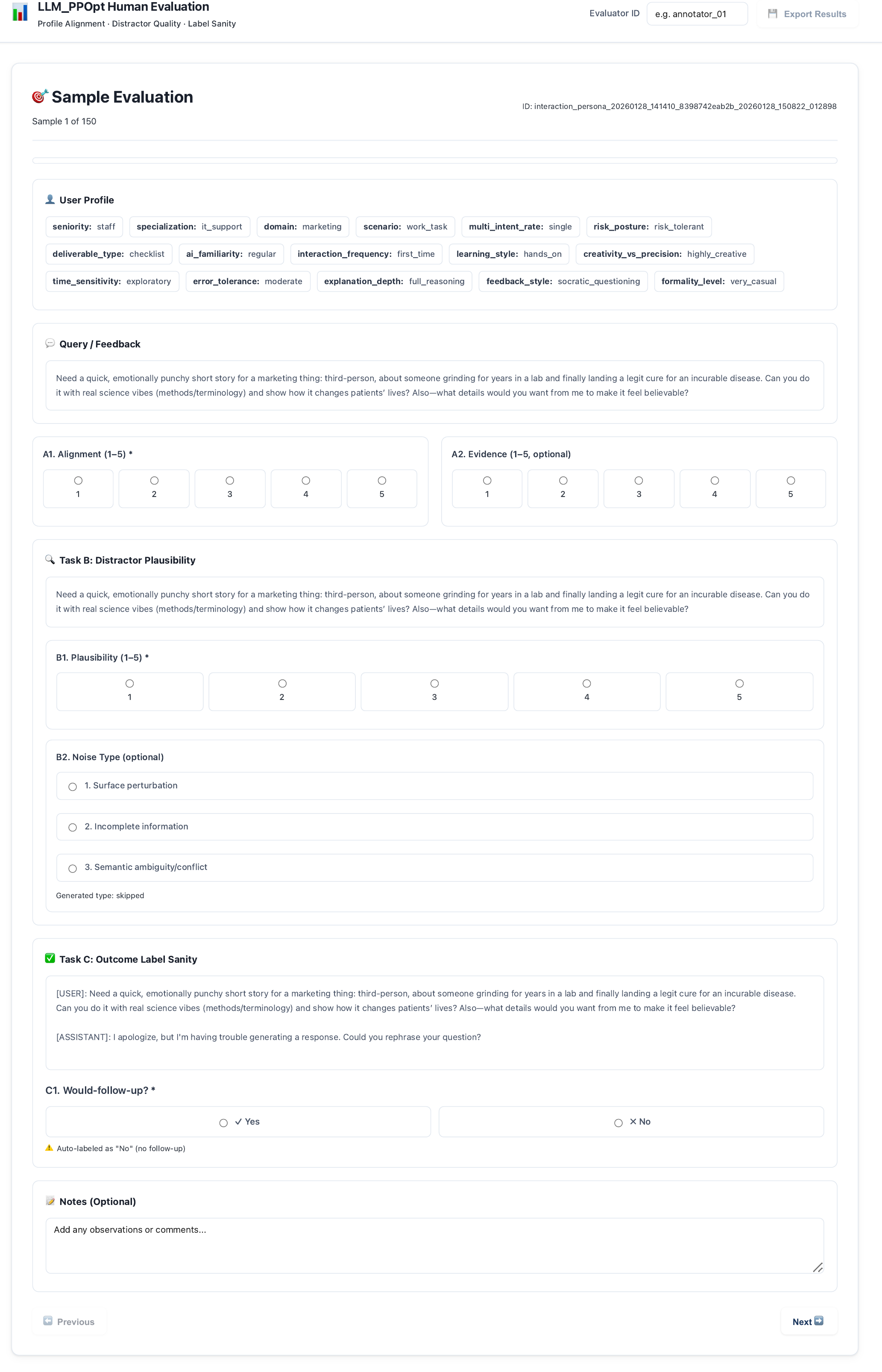}
     \vspace{15pt}
    \caption{Human evaluation interface for the quality analysis of synthetic data via \dataGeneration.}
    \label{fig:placeholder}
\end{figure}

\clearpage


\end{document}